%% file: main_arxiv.tex
\documentclass[10pt,twocolumn,letterpaper]{article}
\usepackage{cvpr}

\def\confName{arXiv}
\def\confYear{2026}

\makeatletter
\newcommand{\appendixtoc}{%
  \begingroup
  \setcounter{tocdepth}{2}
  \@starttoc{apx}%
  \endgroup
}
\makeatother
\usepackage{amsthm}
\newtheorem{theorem}{Theorem}
\newtheorem{proposition}{Proposition}
\usepackage{eccvabbrv}
\usepackage{wrapfig}
\usepackage{graphicx}
\usepackage{booktabs}
\usepackage[table]{xcolor}
\usepackage{colortbl}
\usepackage{array}
\usepackage{multirow}
\usepackage{makecell}
\usepackage{adjustbox}
\usepackage{enumitem}
\usepackage[normalem]{ulem}
\usepackage{tikz}

\usepackage{CJKutf8}

\renewcommand{\multirowsetup}{\centering}
\definecolor{mygray}{gray}{.92}
\definecolor{mygreen1}{RGB}{253, 244, 244}
\definecolor{mygreen2}{RGB}{238, 243, 243}
\definecolor{ForestGreen}{RGB}{34,139,34}
\providecommand{\mathcolor}[2]{\textcolor{#1}{#2}}
\newcommand{\fg}[1]{\mathbf{\mathcolor{ForestGreen}{#1}}}
\definecolor{Forestred}{RGB}{220,50,50}

\usepackage[accsupp]{axessibility}  

\usepackage{hyperref}
\hypersetup{
    colorlinks=true,
    urlcolor=magenta, 
    linkcolor=blue,
    citecolor=green
}
\usepackage{orcidlink}

\newcommand{\ours}{CLSE\xspace}

\begin{document}

\title{Spectral Evolution-Guided Token Pruning in \\ Multimodal Large Language Models} 

\author{Bin Chen$^{1,2}$ \quad Yuxiang Cai$^{1,2}$* \quad
Yadan Luo$^{3}$ \quad Yi Zhang $^{4}$  \quad Jianwei Yin$^{1,2}$  \quad 
Zhi Chen$^{5}$ \thanks{Corresponding Authors: Yuxiang Cai, Zhi Chen} \\
$^1$ School of Software Technology, Zhejiang University, Ningbo, China\\
$^2$ Zhejiang Key Laboratory of Digital-Intelligence Service Technology, China \\
$^3$ The University of Queensland, St Lucia, QLD, Australia \\
$^4$ Singapore Management University, Singapore \\
$^5$ The University of Southern Queensland, Toowoomba, QLD, Australia  \\
\textit{caiyuxiang@zju.edu.cn, uqzhichen@gmail.com} \\
\url{https://github.com/zjubinchen/CLSE}}

\maketitle

\begin{abstract}
Reducing visual token redundancy is critical for accelerating Multimodal Large Language Models (MLLMs) without degrading cross-modal reasoning performance. Existing token pruning methods typically rely on single-layer signals, such as attention scores or token similarities, which overlook the cross-layer transformation of visual representations and may exhibit positional bias in multimodal token sequences. To address this limitation, we propose a training-free token pruning framework based on Cross-Layer Spectral Evolution (CLSE). Instead of measuring token importance from single-layer feature magnitudes, CLSE quantifies how token representations evolve across Transformer layers in the frequency domain. This evolution reflects the transition from high-frequency structural details to low-frequency semantic abstractions. We observe that tokens with stronger spectral redistribution across layers are more likely to be semantically active and should therefore be preserved. By modeling cross-layer token dynamics, CLSE provides a stable importance criterion that mitigates positional bias. Extensive experiments on both image and video benchmarks demonstrate that CLSE achieves a superior trade-off between efficiency and accuracy under aggressive token reduction. Across multiple MLLMs, CLSE reduces FLOPs, KV cache memory, and latency while maintaining competitive or improved performance. 
\end{abstract}

\section{Introduction}
\label{sec:intro}
Multimodal Large Language Models (MLLM)~\cite{liu2023llava,yang2025qwen3} have achieved remarkable progress in visual question answering, grounded dialogue, and multimodal reasoning. 
By coupling a vision encoder (\textit{e.g.,} ViT~\cite{dosovitskiy2020image}) with a powerful LLM, these models enable strong cross-modal alignment and reasoning capabilities. However, this performance comes at a significant computational cost. Modern LLMs process hundreds to thousands of visual tokens per sample, leading to substantial inference latency, particularly in high-resolution and video scenarios.

\begin{figure*}[t]
    \centering
    \includegraphics[width=1\linewidth]{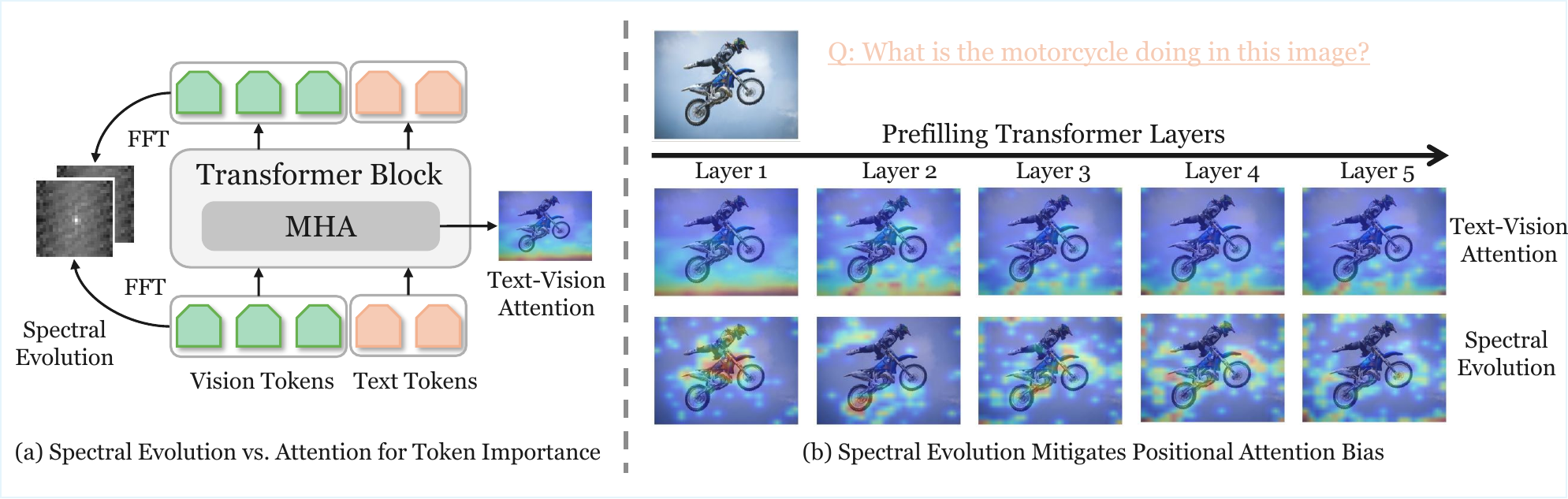}
    \caption{Spectral evolution provides a more reliable token importance signal than attention.
(a) Comparison between conventional text–vision attention and the proposed spectral evolution for estimating visual token importance across a Transformer block. (b) Layer-wise visualization shows that the attention is biased toward the visual tokens appearing later in the flattened sequence due to positional encoding effects, whereas spectral evolution consistently highlights semantically evolving regions across layers.}
    \label{fig:intro}
\end{figure*}

A key source of inefficiency lies in visual token redundancy. Spatial patches in images and even temporal patches in videos exhibit strong local and cross-frame correlations. Yet standard Transformer architectures process all tokens uniformly through deep stacks of self-attention layers. To mitigate this inefficiency, recent works propose dynamic token reduction strategies. For example, PriorTR~\cite{chen2026PriorTRimage}, DynamicViT~\cite{rao2021dynamicvit}, A-ViT~\cite{yin2022vit}, and ATS~\cite{fayyaz2022adaptive} introduce adaptive token pruning mechanisms in ViTs. In the multimodal domain, FastV~\cite{chen2024image} prunes visual tokens after early layers of MLLMs, while SparseVLM~\cite{zhangsparsevlm} adopts progressive sparsification with token recycling. Complementary approaches such as ToMe~\cite{bolyatoken} and PruMerge~\cite{shang2025llava} aggregate similar tokens instead of discarding them, preserving complementary information under aggressive compression.



Despite the effectiveness, most existing methods rely on \emph{instantaneous} signals computed within a single layer, such as attention weights, feature magnitudes, or similarity scores. However, attention is not always a reliable indicator of semantic contribution. As illustrated in Fig.~\ref{fig:intro}, attention scores can exhibit strong positional bias, \textit{i.e.,} concentrating on visual tokens appearing later in the flattened sequence, even when those regions are not semantically critical~\cite{wen2025token}. 
Furthermore, static magnitude-based methods overlook a fundamental characteristic of LLM decoding behavior in an MLLM, where visual token representations undergo structured evolution across LLM layers, transitioning from low-level spatial details to high-level semantic abstractions while the text instruction is fused. This cross-layer transformation, rather than instantaneous attention intensity, more faithfully reflects a token's true contribution to multimodal reasoning.

Recent studies provide a complementary perspective by analyzing Transformers in the frequency domain. Works such as FNet~\cite{lee2022fnet} and Fourier-based vision architectures~\cite{nguyen2022fourierformer} demonstrate that spectral operations can effectively model global dependencies. More recently, frequency-aware compression strategies have been explored for multimodal models~\cite{wang2025fourier}, highlighting the role of low-frequency components in semantic representation. Empirical observations suggest that early Transformer layers preserve high-frequency spatial information, while deeper layers progressively emphasize low-frequency semantic structures. This depth-wise spectral shift reflects the abstraction process underlying multimodal reasoning.

Motivated by these observations, we propose a cross-layer spectral evolution method for visual token reduction in MLLMs, as shown in Fig. \ref{fig:intro} (a). Our key insight is that token importance should be measured by how its representation \emph{evolves} across layers. The tokens that contribute meaningfully to text-vision reasoning show stronger spectral evolution during the detail-to-semantics transformation, whereas redundant background tokens remain relatively stationary in the frequency domain. 
Specifically, we  
(1) project visual tokens into the frequency domain and apply a Gaussian high-pass filter to isolate global information;  
(2) quantify cross-layer spectral evolution as a token-wise transformation intensity; and  
(3) integrate this evolution signal to obtain a robust importance estimate.
Unlike purely attention-driven or magnitude-based pruning strategies, our approach captures \emph{representation dynamics} across layers. Furthermore, our framework is orthogonal to token merging methods and can be seamlessly combined with merging strategies such as ToMe~\cite{bolyatoken}. This complementarity is particularly beneficial in video settings, where temporal redundancy amplifies token correlations.
Extensive experiments on image and video benchmarks demonstrate that \ours consistently improves the trade-off between efficiency and accuracy across multiple MLLMs. 
In summary, our contributions are threefold:
\begin{itemize}
\item We introduce the cross-layer spectral evolution framework, dubbed CLSE, as a token importance estimation principle for MLLMs, which characterizes how visual representations evolve from structural details to semantic abstractions across LLM layers.
\item We design a {training-free spectral-guided token pruning framework} that operationalizes CLSE to identify and remove redundant tokens during inference while preserving tokens undergoing meaningful representational evolution.
\item We conduct extensive experiments on both image- and video-based multimodal benchmarks, showing that CLSE achieves a superior trade-off between efficiency and accuracy under aggressive token reduction. Across multiple MLLMs, our method consistently reduces FLOPs, KV cache memory, and latency while maintaining competitive or improved performance.
\end{itemize}

\section{Related Work}

\noindent \textbf{Training-Free Token Reduction for MLLMs.}
Multimodal Large Language Models (MLLMs)~\cite{yang2025qwen3,liu2024improved,chen2024internvl,liu2023llava} incur substantial overhead due to long visual token sequences. 
In applications where efficiency matters \cite{chen2025fastedit, wang2026learning, chen2026distributed, chen2025cluster, chen2025svip,chen2023zero,chen2022gsmflow,chen2021semantics,chen2021mitigating,zhao2026generative,shao2026tr,li2025pataug,wang2025discrimination,zhang2024towards,zhao2025continual,yu2025dynamic,zhao2025synthetic,lim2024dipex,you2022pixel,su2022distinguishing,you2021domain,chen2020rethinking,chen2020canzsl}, reducing the number of visual tokens during inference has emerged as an effective strategy for accelerating MLLMs without retraining. Existing training-free methods typically estimate token importance using heuristics derived from intermediate representations and can be broadly categorized into attention-based, similarity-based, diversity-based, and hybrid approaches.
\textit{Attention-based methods} estimate token importance directly from cross-modal attention between text and visual tokens.
FastV~\cite{chen2024image} is a representative plug-and-play approach that ranks visual tokens using attention scores at a selected transformer layer and prunes tokens with the lowest attention magnitude for subsequent layers.
Several follow-up works extend this idea using different attention aggregation or dynamic selection strategies~\cite{zeng2025glimpse,xing2025vision,ye2025fit,sun2026ivc,song2025less,wang2025corematching,liu2024prompt,he2024zipvl,han2024rethinking,Zhuang_2025_CVPR,liang2025dynamic}.
These approaches are computationally efficient and easy to integrate with existing MLLMs, but they generally rely on absolute attention magnitude as the ranking signal.
\textit{Similarity-based methods} identify redundant visual tokens by measuring similarity in feature space and removing tokens that provide overlapping information.
For example, DART~\cite{wen2025stop} detects redundant visual tokens through similarity-based matching and selectively retains a compact subset of representative tokens. 
Related approaches also employ clustering or feature distance criteria~\cite{jeddi2025similarity,huang2024ivtp,yang2025topv,siinfoprune,yu2026visiontrim}.
\textit{Diversity-based methods} aim to maintain a representative set of visual tokens by maximizing feature diversity among the selected tokens.
For instance, DivPrune~\cite{alvar2025divprune} encourages diversity in the retained tokens to avoid redundancy and improve coverage of different visual regions.
Other approaches formulate token selection as an entropy or diversity maximization problem to preserve informative structures in the visual representation~\cite{wang2026entropyprune,zhang2025beyond,zamini2026delta}.
While diversity criteria help prevent redundancy, they may not explicitly capture task relevance.
\textit{Hybrid methods} combine multiple criteria to improve token selection robustness.
VisPruner~\cite{fan2025visipruner} integrates attention-based importance with diversity-aware selection, while ToDre~\cite{li2025todre} jointly considers diversity and attention signals when selecting tokens.
\textit{Token merging approaches} reduce sequence length by merging visually similar tokens rather than removing them.
Representative methods include LightVLM~\cite{hu2025lightvlm}, CrossGet~\cite{shi2023crossget}, VisionZip~\cite{yang2025visionzip}, and SparseVLM~\cite{zhangsparsevlm}.
A concurrent work V²Drop~\cite{chen2026variation} explores intrinsic cross-layer dynamics by progressively dropping 'lazy tokens' with minimal representational variation.
In contrast to existing methods that rely on instantaneous signals, we propose to measure \emph{cross-layer spectral evolution} of tokens. By analyzing how token representations evolve across layers in the frequency domain, our approach captures structural-to-semantic transitions that are not visible from single-layer statistics, leading to more reliable identification of informative visual tokens.

\noindent \textbf{Frequency Domain in Transformers.}
A growing body of work interprets Transformers through a frequency-domain lens. In language, FNet~\cite{lee2022fnet} replaces self-attention with Fourier mixing to accelerate computation, demonstrating that spectral operations can serve as efficient token mixers. VTC-LFC~\cite{wang2022vtc} compresses Vision Transformers by retaining only low-frequency spectral components of token representations.
SFHformer~\cite{jiang2024fast} introduces frequency-domain token mixing via FFT to model global interactions with lower computational cost than standard self-attention, demonstrating the effectiveness of spectral operations in vision Transformers.
In multimodal models, frequency-domain compression has also been explored; for instance, Fourier-VLM \cite{wang2025fourier} compresses vision tokens by applying low-pass filtering in the frequency domain. Different from these approaches that apply spectral transforms for compression within a layer, we use cross-layer spectral evolution as an importance criterion, capturing whether a token undergoes meaningful detail-to-semantics transformation across layers.


\begin{figure*}[t]
    \centering
    \includegraphics[width=1\linewidth]{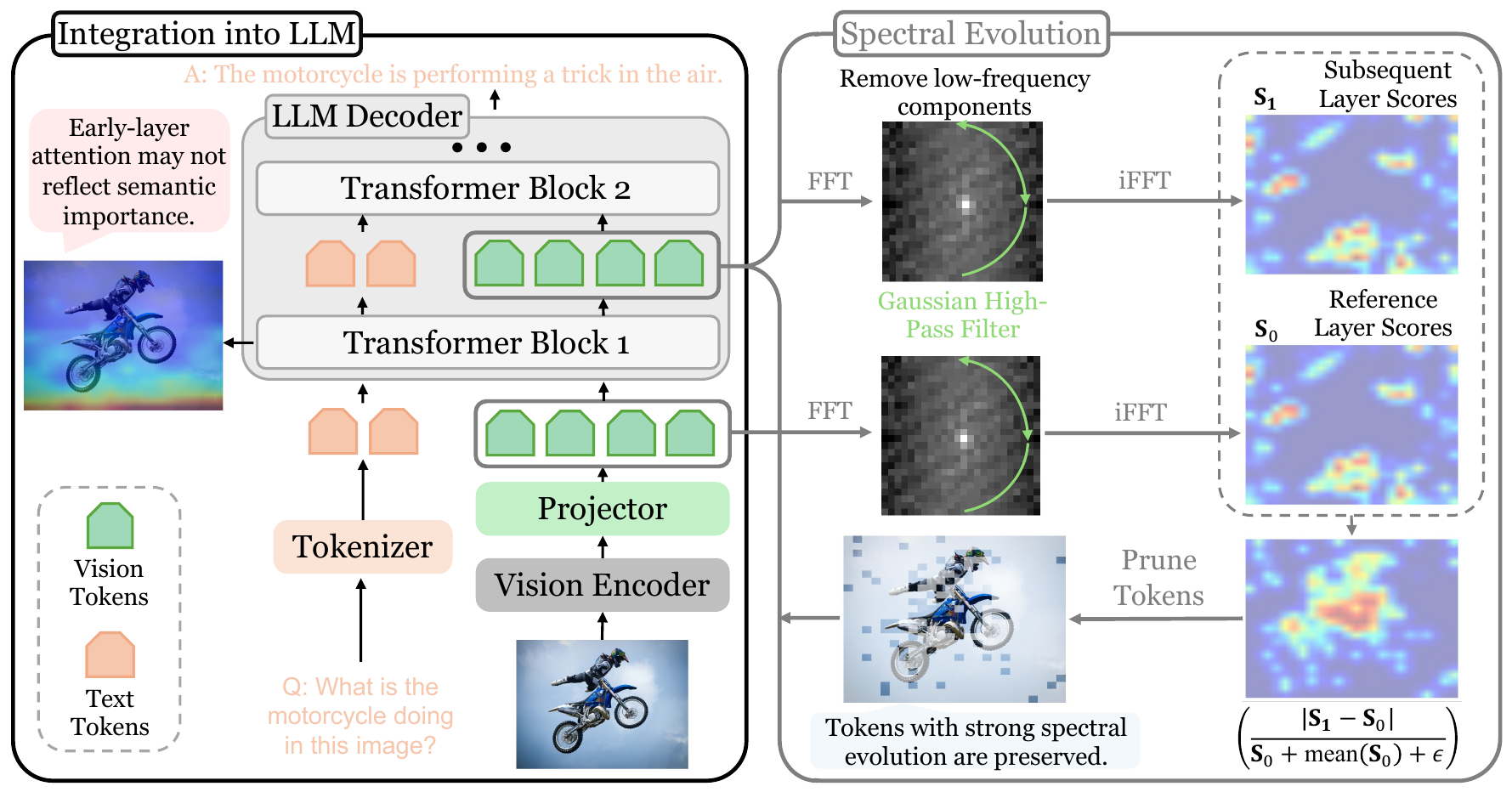}
    \caption{Overview of the proposed spectral evolution-guided token pruning framework. \textbf{Left}: Cross-layer spectral evolution is computed by transforming layer-wise token representations into the frequency domain, suppressing low-frequency components via a Gaussian high-pass filter, and measuring the normalized spectral difference between adjacent layers. Tokens exhibiting stronger spectral redistribution are considered decision-critical and preserved. \textbf{Right}: The \ours  module is integrated into early LLM decoder layers to guide token selection. Compared with raw attention scores, CLSE provides a more reliable signal for identifying semantically evolving tokens.}
    \label{fig:framework}
\end{figure*}

\section{Method}
\subsection{Preliminaries}
\label{preliminary}
\noindent \textbf{Multimodal Large Language Models.}
An MLLM consists of a vision encoder $f_v$, a LLM decoder $f_l$, and a projection module $\phi(\cdot)$ that maps visual embeddings into the LLM token space.
Given an image or video frames $I$ and text prompt $T$, we obtain:
\begin{equation}
\mathbf{X}^0 = \phi(f_v(I)) \in \mathbb{R}^{N \times d},\qquad
\mathbf{Z}^0 = \mathrm{Embed}(T) \in \mathbb{R}^{M \times d},
\end{equation}
where $N$ and $M$ are the numbers of visual and text tokens and $d$ is the shared embedding dimension.
We concatenate tokens into a joint sequence processed by $L$ Transformer layers:
\begin{align}
\mathbf{H}^0 &= [\mathbf{Z}^0; \mathbf{X}^0] \in \mathbb{R}^{(M+N)\times d},\qquad \\
\mathbf{H}^{\ell}&=\mathrm{Transformer}^{\ell}(\mathbf{H}^{\ell-1}),\ \ \ell=1,\dots,L.
\end{align}
For convenience, we denote the \emph{visual slice} of layer $\ell$ as
$\mathbf{X}^{\ell} \in \mathbb{R}^{N\times d}$.

\noindent \textbf{Token Redundancy and Reduction.}
Visual tokens typically exhibit strong spatial and semantic redundancy. Not all tokens contribute equally to cross-modal reasoning: background regions or visually repetitive patches may have limited influence on the final output. Token reduction aims to select a subset of $K \ll N$ informative tokens:
\begin{equation}
\tilde{\mathbf{X}} = \mathcal{S}(\mathbf{X}),
\quad
\tilde{\mathbf{X}} \in \mathbb{R}^{K \times d},
\end{equation}
where $\mathcal{S}$ is a selection operator that preserves tokens with high semantic contribution while discarding redundant ones. After selection, the reduced multimodal sequence becomes:
\begin{equation}
\tilde{\mathbf{H}} = 
\left[ \mathbf{Z} ; \tilde{\mathbf{X}} \right],
\end{equation}
which is then processed by subsequent Transformer layers.

\noindent \textbf{Fast Fourier Transform (FFT).}
Let $\mathbf{X}\in\mathbb{R}^{H\times W}$ denote a 2D signal, where $H$ and $W$ are spatial height and width, and $X_{h,w}$ is the value at coordinate $(h,w)$ with $h \in \{0,\dots,H-1\}$ and $w \in \{0,\dots,W-1\}$. 
The 2D Discrete Fourier Transform (DFT) maps $\mathbf{X}$ to a frequency representation $\hat{\mathbf{X}} \in \mathbb{C}^{H \times W}$:
\begin{equation}
\hat{X}_{u,v}
=
\sum_{h=0}^{H-1}
\sum_{w=0}^{W-1}
X_{h,w}
e^{\!
- j 2\pi
\left(
\frac{uh}{H} + \frac{vw}{W}
\right)},
\end{equation}
where $(u,v)$ are frequency indices and $j=\sqrt{-1}$. We denote the forward and inverse transforms as $\hat{\mathbf{X}} = \mathcal{F}(\mathbf{X})$ and $\mathbf{X} = \mathcal{F}^{-1}(\hat{\mathbf{X}})$, respectively. The DFT decomposes $\mathbf{X}$ into orthogonal complex exponentials of increasing spatial frequency: low-frequency coefficients encode global, slowly varying structures, while high-frequency coefficients capture fine-grained spatial variations.

\subsection{Cross-Layer Spectral Evolution (\ours )}
In Fig. \ref{fig:framework}, we present an intuitive overview of our CLSE.
Let $N=H\cdot W$ and reshape the visual tokens $\mathbf{X}^{\ell}\in\mathbb{R}^{N\times d}$ into spatial feature maps $\mathbf{X}^{\ell}_{2\mathrm{D}}\in\mathbb{R}^{H\times W \times d}$, we measure high-frequency structural energy instead of raw embedding magnitude. We first transform to the frequency domain and centre the spectrum:
\begin{equation}
\widehat{\mathbf{X}}^{\ell}=\mathrm{fftshift}(\mathcal{F}(\mathbf{X}^{\ell}_{2\mathrm{D}})).
\end{equation}
To suppress low-frequency global patterns and highlight local structure, we apply a Gaussian high-pass mask
$\mathbf{M}\in\mathbb{R}^{H\times W}$:
\begin{equation}
\mathbf{M}_{u,v}=1-\exp\!\left(
-\frac{(u-c_u)^2+(v-c_v)^2}{2(r\cdot c_u)^2}
\right),
\end{equation}
where $(c_u,c_v)$ is the spectral centre and $r$ is the cutoff ratio.
We then invert the transform to obtain a spatial residual map and average over channels:
\begin{align}
\mathbf{R}_{\ell} &=
\left|\mathcal{F}^{-1}\!\left(\mathrm{ifftshift}(\widehat{\mathbf{X}}^{\ell}\odot \mathbf{M})\right)\right|,
\qquad \\
\mathbf{S}_{\ell}&=\mathrm{Avg}_d(\mathbf{R}_{\ell})\in\mathbb{R}^{H\times W}.
\end{align}
Flattening $\mathbf{S}_{\ell}$ over spatial positions yields $\mathbf{S}_{\ell}\in\mathbb{R}^{N}$,
a per-token distribution of high-frequency energy. Crucially, $\mathbf{S}_{\ell}$ is \emph{not}
the full feature norm. Instead, it isolates band-limited residual energy after low-frequency suppression, where an ablation study evidenced this in Fig. \ref{fig:ablation}(c).

\ours measures how a token's band-limited structure changes across depth throughout the process of text-vision interaction in the transformer blocks. Background tokens tend to exhibit near-static residual patterns, while semantically engaged tokens often show pronounced redistribution.
For a reference layer $\ell$ and a subsequent layer $\ell{+}1$, we define the {evolution intensity}:
\begin{align}
\mathcal{I}^{(\ell)} &=
\mathrm{clip}\!\left(
\frac{|\mathbf{S}_{\ell+1}-\mathbf{S}_{\ell}|}{
\mathbf{S}_{\ell}+ \overline{\mathbf{S}}_{\ell} + \epsilon},
\ 0,\ \tau_{\max}
\right),
\qquad \\
\overline{\mathbf{S}}_{\ell}&=\mathrm{mean}(\mathbf{S}_{\ell}\ \text{over tokens}),
\end{align}
where $\mathcal{I}^{(\ell)}$ is a relative metric to achieve scale-invariance. Dividing by $\mathbf{S}_{\ell}$ prevents tokens with large embedding magnitudes from dominating the signal, while the global mean $\overline{\mathbf{S}}_{\ell}$ acts as a regularizer to further suppress minor spectral fluctuations in inactive or redundant regions (where local energy is below the image-wide average). $\epsilon$ and $\tau_{\max}$ ensure numerical stability and outlier robustness. 

To maximize prefilling speedups, we perform early-stage pruning in the LLM decoder. In our default setting, we compute \ours from layer $0\!\rightarrow\!1$, which is between input embeddings $\mathbf{X}^0$ and the first decoder layer output $\mathbf{X}^1$, and execute pruning before layer $2$. For each sample, we select the top-$K$ tokens based on the evolution intensities:
\begin{equation}
\mathcal{K}=\mathrm{TopK}({\mathcal{I}}^{(\ell)},K),\qquad\tilde{\mathbf{X}}^{\ell}=\mathbf{X}^{\ell}[\mathcal{K}]\in\mathbb{R}^{K\times d},\end{equation}and continue decoding with the truncated sequence $\tilde{\mathbf{H}}^{\ell}=[\mathbf{Z}^{\ell};\tilde{\mathbf{X}}^{\ell}]$ for all subsequent layers. This yields a training-free, plug-and-play pruning module that significantly reduces computational overhead.

\subsection{Theoretical Analysis}
\label{sec:theory}
The above pruning rule admits a simple fidelity interpretation. Specifically, the calibrated spectral evolution score ${\mathcal I}^{(\ell)}$
quantifies the amount of cross-layer band-limited structural redistribution associated with each token. For any retained token set $\Omega$ with $|\Omega|=K$, let $P_\Omega(\mathbf X^\ell)$ denote the pruned token sequence. Then the pruning-induced perturbation of the subsequent decoder $g(\cdot)$ admits the bound:
\begin{equation}
\|g(\mathbf X^\ell)-g(P_\Omega(\mathbf X^\ell))\|_2
\le
\alpha \sum_{i\notin\Omega} {\mathcal I}^{(\ell)}_i ,
\end{equation}
where the right-hand side corresponds to the discarded spectral evolution mass. Therefore, selecting the top-$K$ tokens by CLSE minimizes this upper bound
among all subsets of size $K$, \textit{i.e.,} it preserves the maximum amount of
cross-layer spectral evolution under a fixed token budget.
A derivation is provided in the appendix.

\begin{table*}[t]
	\caption{Performance comparison with baselines on nine image understanding benchmarks with LLaVA-1.5-7B. The vanilla 576 tokens are pruned to 192, 128, and 64.}
    \centering
    \centering
    \begin{adjustbox}{width=0.96\textwidth,center}
    \begin{tabular}{l | ccccccccc | >{\centering\arraybackslash}p{1.0cm}}
        \toprule[1.5pt]
        \textbf{Method} & \textbf{GQA} & \textbf{MMB} & \textbf{MMB}$_{CN}$ & \textbf{MME} & \textbf{POPE} & \textbf{SQA} &  \textbf{VQA}$_{\text{Text}}$ & \textbf{VizWiz} & \textbf{OCR}$_{\text{Bench}}$ & \makecell[c]{\textbf{Avg}.}\\
        \hline
        \rowcolor{mygray}
        LLaVA-1.5-7B & \multicolumn{10}{c}{\textit{Upper Bound, 576 Tokens} \ $\textbf{(100\%)}$}\\
        \textcolor{gray}{Vanilla} & \textcolor{gray}{61.9} & \textcolor{gray}{64.7} & \textcolor{gray}{58.1} & \textcolor{gray}{1862} & \textcolor{gray}{85.9} & \textcolor{gray}{69.5}  & \textcolor{gray}{58.2} & \textcolor{gray}{50.0} & \textcolor{gray}{29.7} & \multirow{1}*{\textcolor{gray}{100.0\%}} \\

        \midrule
        \rowcolor{mygray}
         & \multicolumn{10}{c}{\textit{Retain 192 Tokens} \ $\fg{(\downarrow 66.7\%)}$}\\
        ToMe \texttt{\scriptsize{(ICLR23)}} & 54.3 & 60.5 & 53.2 & 1563  & 65.2 & 68.0 & 52.1 & 50.9 & 25.8 & \multirow{1}*{84.8\%} \\
        FastV \texttt{\scriptsize{(ECCV24)}} & 52.7 & 61.2 & 57.0 & 1612  & 67.3 & 67.1 & 52.5 & 50.8 & 29.1 & \multirow{1}*{92.1\%} \\
        PruMerge \texttt{\scriptsize{(ICCV25)}}\;\; & 54.3 & 59.6 & 52.9 & 1632 & 71.3 & 67.9  & 54.3 & 50.1 &25.3 & 90.8\% \\
        PDrop \texttt{\scriptsize{(CVPR25)}} & 57.1 & 63.2 & 56.8 & 1766 & 82.3 & 68.8  & 56.1 & \textcolor{MidnightBlue}{\textbf{51.1}} &28.7 & 96.9\% \\
        HiRED \texttt{\scriptsize{(AAAI25)}} & 58.7 & 62.8 & 54.7 & 1737 & 82.8 & 68.4  & 47.4 & 50.1 & 19.0 & 91.0\%     \\
        SparseVLM \texttt{\scriptsize{(ICML25)}} & 57.6 & 62.5 & 53.7 & 1721 & 83.6 & 69.1  & 56.1 & 50.5 &29.2 & 96.3\% \\
        FiCoCo-V \texttt{\scriptsize{(AAAI26)}} & 58.5 & 62.3 & 55.3 & 1732 & 82.5 & 67.8  & 55.7 & 51.0 & 28.6 & 96.2\% \\
        \rowcolor{cyan!7}
        \ours \scriptsize{(Ours)} & \textcolor{MidnightBlue}{\textbf{61.5}} & \textcolor{MidnightBlue}{\textbf{63.6}} & \textcolor{MidnightBlue}{\textbf{57.2}} & 
        \textcolor{MidnightBlue}{\textbf{1817}} & \textcolor{MidnightBlue}{\textbf{85.1}} & \textcolor{MidnightBlue}{\textbf{70.1}} & \textcolor{MidnightBlue}{\textbf{57.5}} & 50.6 & \textcolor{MidnightBlue}{\textbf{30.1}}  & \textcolor{MidnightBlue}{\textbf{99.4\%}} \\
        \midrule
         \rowcolor{mygray}
         & \multicolumn{10}{c}{\textit{Retain 128 Tokens} \ $\fg{(\downarrow 77.8\%)}$}\\
        ToMe \texttt{\scriptsize{(ICLR23)}} & 52.4 & 53.3 & 50.7 & 1343 & 62.8 & 59.6 & 49.1 & \textcolor{MidnightBlue}{\textbf{51.6}} & 25.0 & \multirow{1}*{84.1\%} \\
        FastV \texttt{\scriptsize{(ECCV24)}} & 49.6 & 56.1 & 56.4 & 1490 & 59.6 & 60.2 & 50.6 & 51.3  & 28.5 & \multirow{1}*{87.2\%}\\
        PruMerge \texttt{\scriptsize{(ICCV25)}} & 53.3 & 58.1 & 51.7 & 1554 & 67.2 & 67.1 & 54.3 & 50.3 &24.8 & 88.9\%  \\
        PDrop \texttt{\scriptsize{(CVPR25)}} & 56.0 & 61.1 & \textcolor{MidnightBlue}{\textbf{56.6}} & 1644 & {82.3} & 68.3  & 55.1 & 51.0 & \textcolor{MidnightBlue}{\textbf{28.7}} &95.3\% \\
        HiRED \texttt{\scriptsize{(AAAI25)}} & 57.2 & 61.5 & 53.6 & 1710 & 79.8 & 68.1  & 46.1 & 51.3 & 19.1 & 89.8\% \\
        SparseVLM\texttt{\scriptsize{(ICML25)}} & 56.0 & 60.0 & 51.1 & 1696 & 80.5 & 67.1 & 54.9 & 51.4 & 28.0 &93.7\% \\
        FiCoCo-V \texttt{\scriptsize{(AAAI26)}} & 57.6 & 61.1 & 54.3 & 1711 & 82.2 & 68.3 & 55.6 & 49.4 & 26.2 & 94.3\% \\

        \rowcolor{cyan!7} \ours \scriptsize{(Ours)} & \textcolor{MidnightBlue}{\textbf{60.9}} & 
        \textcolor{MidnightBlue}{\textbf{63.0}} & {{55.8}} & \textcolor{MidnightBlue}{\textbf{1816}} & \textcolor{MidnightBlue}{\textbf{84.4}} & \textcolor{MidnightBlue}{\textbf{70.1}}  & \textcolor{MidnightBlue}{\textbf{56.6}} & 51.0  & 28.4 & \textcolor{MidnightBlue}{\textbf{98.1\%}} \\
        \midrule

       \rowcolor{mygray}
         & \multicolumn{10}{c}{\textit{Retain 64 Tokens} \ $\fg{(\downarrow 88.9\%)}$}\\
        ToMe \texttt{\scriptsize{(ICLR23)}} & 48.6 & 43.7 & 47.1 & 1138 & 52.5 & 50.0  & 45.3 & \textcolor{MidnightBlue}{\textbf{52.6}}  & 22.7 & \multirow{1}*{67.4\%}\\
        FastV \texttt{\scriptsize{(ECCV24)}} & 46.1 & 48.0 & 52.7 & 1256 & 48.0 & 51.1  & 47.8 & 50.8 & 24.5 & 78.0\% \\
        PruMerge \texttt{\scriptsize{(ICCV25)}} & 51.9 & 55.3 & 49.1 & 1549 & 65.3 & 68.1 & 54.0 & 50.1 & 25.0 & 87.5\% \\
        PDrop \texttt{\scriptsize{(CVPR25)}} & 41.9 & 33.3 & 50.5 & 1092 & 55.9 & 68.6  & 45.9 & 50.7 & 25.0 & 77.0\% \\
        HiRED \texttt{\scriptsize{(AAAI25)}} & 54.6 & 60.2 & 51.4 & 1599 & 73.6 & 68.2  & 44.2 & 50.2 & 19.1 & 86.6\% \\
        SparseVLM \texttt{\scriptsize{(ICML25)}} & 52.7 & 56.2 & 46.1 & 1505 & 75.1 & 62.2  & 51.8 & 50.1 & 18.0 & 84.3\% \\
        FiCoCo-V \texttt{\scriptsize{(AAAI26)}} & 52.4 & 60.3 & 53.0 & 1591 & 76.0 & 68.1  & 53.6 & 49.8 &22.6 & 89.8\% \\
      
        \rowcolor{cyan!7} \ours \scriptsize{(Ours)} & \textcolor{MidnightBlue}{\textbf{58.3}} & \textcolor{MidnightBlue}{\textbf{61.9}} & \textcolor{MidnightBlue}{\textbf{54.3}}  & \textcolor{MidnightBlue}{\textbf{1809}} & \textcolor{MidnightBlue}{\textbf{78.3}} & 
        \textcolor{MidnightBlue}{\textbf{69.6}} & \textcolor{MidnightBlue}{\textbf{55.5}}  & 
        50.9 & 
        \textcolor{MidnightBlue}{\textbf{25.1}} & \textcolor{MidnightBlue}{\textbf{94.8\%}} \\
        \bottomrule
	\end{tabular}
    \end{adjustbox}
    \label{tab:main_1_5}
\end{table*}

\section{Experiments}
To validate the effectiveness of our \ours framework, we conduct experiments on image and video question-answering datasets with different MLLM variants. 

\subsection{Image VQA Tasks}
\noindent\textbf{Experiment setup.}
We evaluate \ours on nine diverse multimodal benchmarks spanning visual reasoning (GQA~\cite{hudson2019gqa}, SQA~\cite{lu2022learn},
VizWiz~\cite{gurari2018vizwiz}),
hallucination detection (POPE~\cite{li2023evaluating}),
OCR (VQA$_{\text{TEXT}}$~\cite{singh2019towards}, OCR$_{\text{Bench}}$~\cite{liu2024ocrbench}), 
and holistic understanding (MME~\cite{fu2025mme}, MMB~\cite{liu2024mmbench}, MMB$_{CN}$~\cite{liu2024mmbench}, MMVet~\cite{yu2024mmvet}).
We compare against seven state-of-the-art training-free methods:
ToMe~\cite{bolyatoken}, FastV~\cite{chen2024image},
PDrop~\cite{xing2024pyramiddrop}, 
SparseVLM~\cite{zhangsparsevlm},
FiCoCo~\cite{FiCoCo2025},
HiRED~\cite{arif2025hired},
and PruMerge~\cite{shang2025llava}.
To assess cross-model generalizability, we evaluate on
LLaVA-1.5-7B~\cite{liu2023llava}, LLaVA-1.5-13B~\cite{liu2023llava}, LLaVA-Next-7B~\cite{liu2024llavanext}, Qwen2-VL-7B~\cite{wang2024qwen2vl}, and Qwen2-VL-72B~\cite{wang2024qwen2vl}. 

\noindent\textbf{Implementation Details.}
We apply \ours to each model's standard inference pipeline without
modifying model weights or requiring additional training.
Unlike other methods that perform pruning in later layers, for example, FastV~\cite{chen2024image} performs at layer 3, \ours performs token pruning at the first layer of the LLM decoder. All experiments are conducted on NVIDIA RTX PRO 6000 GPU.

\noindent\textbf{Main Results.}
Table \ref{tab:main_1_5} reports results on LLaVA-1.5-7B under three token-retention budgets (192, 128, and 64 tokens), corresponding to pruning ratios of 66.7\%, 77.8\%, and 88.9\%, respectively.
\ours applies spectral evolution scoring with simple hard pruning and no token recycling.
Despite this simplicity, \ours achieves the best overall average of 99.4\%, 98.1\%, and 94.8\% across the three budgets, consistently outperforming both hard-pruning methods (FastV, PDrop) and token-merging methods (SparseVLM, PruMerge, FiCoCo-V).
This demonstrates that spectral evolution provides a more reliable token-importance criterion than attention- or magnitude-based strategies, regardless of whether the baseline uses hard pruning or token merging.

\begin{figure*}[t]
  \centering
  \begin{subfigure}[b]{0.33\linewidth}
    \centering
    \includegraphics[width=\linewidth]{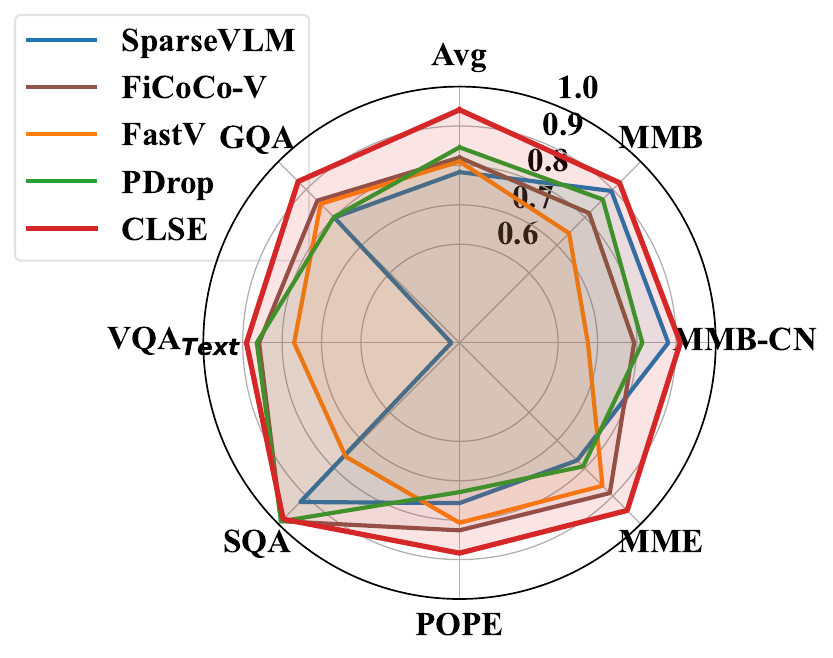}
    \caption{64 Tokens}
    \label{fig:teaser-a}
  \end{subfigure}\hfill
  \begin{subfigure}[b]{0.33\linewidth}
    \centering
    \includegraphics[width=\linewidth]{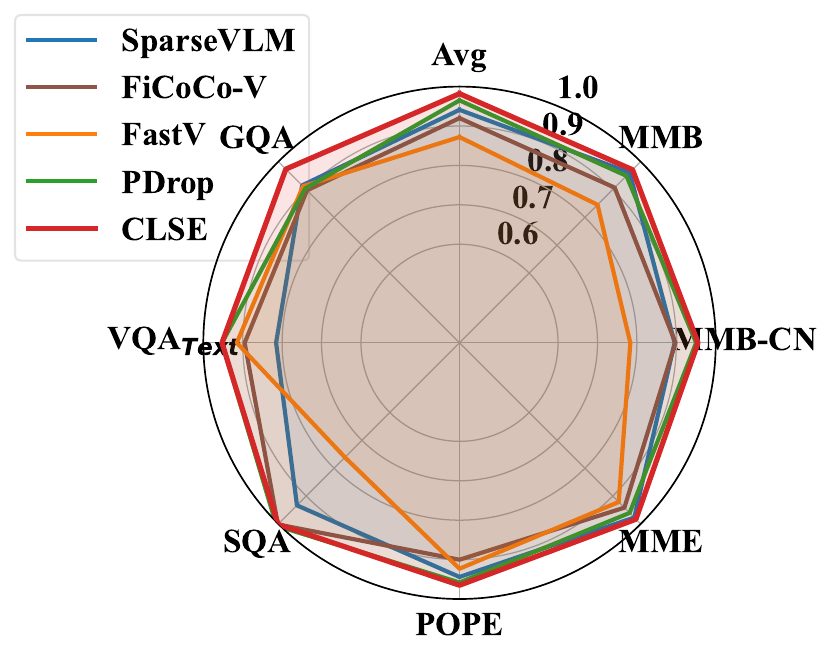}
    \caption{128 Tokens} 
    \label{fig:teaser-b}
  \end{subfigure}\hfill
  \begin{subfigure}[b]{0.33\linewidth}
    \centering
    \includegraphics[width=\linewidth]{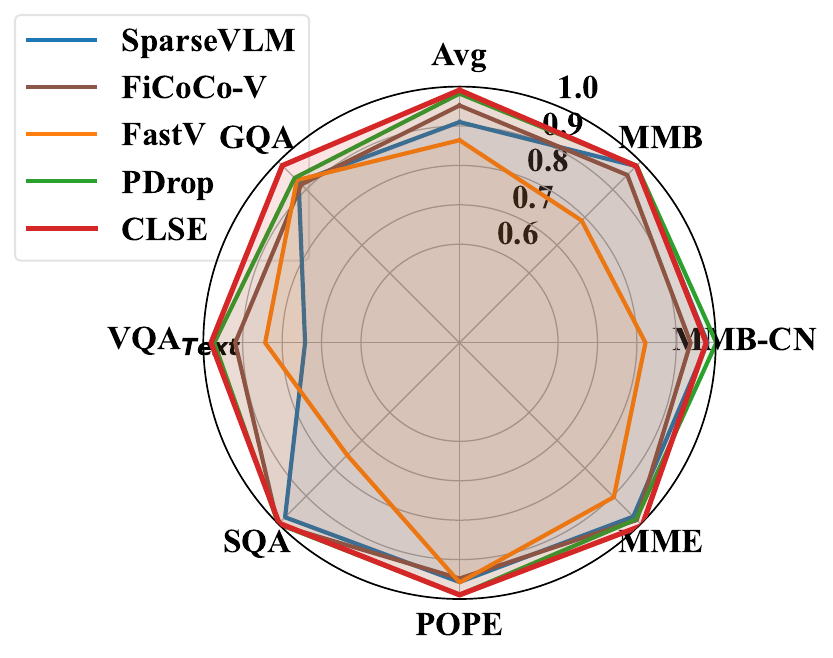}
    \caption{192 Tokens}
    \label{fig:teaser-c}
  \end{subfigure}
  \caption{Radar plots comparing pruning methods on LLaVA-1.5-13B under different token budgets. All metrics, including the overall average (Avg), are normalized to the full-token model.  Across all configurations, \ours achieves the most balanced performance and maintains the highest overall accuracy under aggressive compression. }
  \label{fig:13B}
\end{figure*}

\noindent \textbf{Scaling to Larger Models.}
We further conduct experiments on the larger LLaVA-1.5-13B under $K=192,128,64$ tokens. Figure~\ref{fig:13B} shows that \ours maintains the most balanced performance profile and achieves the strongest overall average across all budgets. 
The advantage becomes more evident as compression increases, where competing methods exhibit larger task-specific degradation. 
These results demonstrate that spectral evolution provides stable token importance estimation and scales effectively to larger models under aggressive pruning. Additionally, we provide an experimental comparison with baseline methods on an even larger model Qwen2-VL-72B in the appendix due to the page limit.

\begin{table*}[t]
	\caption{Performance comparison with baseline methods under different token configurations on Qwen2-VL-7B.}
	\centering
    \resizebox{0.85\textwidth}{!}{%
    \begin{tabular}{p{3.5cm} | c c c c c c c| >{\centering\arraybackslash}p{1.2cm}}
        \toprule[1.5pt]
        \textbf{Method} & \textbf{GQA} & \textbf{MMB} & \textbf{MMB}$_{CN}$  & \textbf{MME} & \textbf{POPE} & \textbf{SQA}& \textbf{VQA}$_{\text{Text}}$ & \makecell[c]{\textbf{Avg}.}\\
        \hline
        \rowcolor{mygray}
        Qwen2-VL-7B & \multicolumn{8}{c}{\textit{Upper Bound, All Tokens} \ $\textbf{(100\%)}$}   \\
        \textcolor{gray}{Vanilla} & \textcolor{gray}{62.2} & \textcolor{gray}{79.1} & \textcolor{gray}{77.8} & \textcolor{gray}{2296} & \textcolor{gray}{87.7} & \textcolor{gray}{85.4} & \textcolor{gray}{81.4} & \multirow{1}*{\textcolor{gray}{100\%}} \\
        \hline
        
        \rowcolor{mygray}
        Qwen2-VL-7B & \multicolumn{8}{c}{\textit{Token Reduction} \ $\fg{(\downarrow 66.7\%)}$}   \\
        \hspace{0.5em} + FastV \texttt{\scriptsize{(ECCV24)}} & 58.6 & 74.5 & 74.3 & 2083 & 85.6 & 83.1 & 76.1 & 94.9\%  \\
        \hspace{0.5em} + PDrop\texttt{\scriptsize{(CVPR25)}} & 58.5 & 72.3 & 71.2 & 1991 & 86.1 & 82.8 & 76.4 & 93.1\%  \\
        \hspace{0.5em} + HiRED \texttt{\scriptsize{(AAAI25)}}  & 59.9 & 73.8 & 73.4 & 2158 & 86.6 & 80.7 & 71.0 & 94.1\% \\
        \hspace{0.5em} + SparseVLM\texttt{\scriptsize{(ICML25)}}  & 59.2 & \textcolor{MidnightBlue}{\textbf{77.1}} & \textcolor{MidnightBlue}{\textbf{76.0}} & 2181 & 85.9 & 83.5 & 77.8 & 96.7\% \\
        \rowcolor{cyan!7}
        \hspace{0.5em} + \ours\texttt{\scriptsize{(ours)}} & \textcolor{MidnightBlue}{\textbf{61.0}} & 76.5 & 75.5 & \textcolor{MidnightBlue}{\textbf{2254}} & \textcolor{MidnightBlue}{\textbf{86.7}} & \textcolor{MidnightBlue}{\textbf{84.2}} & \textcolor{MidnightBlue}{\textbf{78.3}} & \textcolor{MidnightBlue}{\textbf{97.6\%}}  \\
        \hline
    
        \rowcolor{mygray}
        Qwen2-VL-7B & \multicolumn{8}{c}{\textit{Token Reduction} \ $\fg{(\downarrow 77.8\%)}$}\\
        \hspace{0.5em} + FastV \texttt{\scriptsize{(ECCV24)}} & 56.2 & 70.1 & 72.4 & 1957 & 82.7 & 82.3 & 73.0 & 91.3\% \\
        \hspace{0.5em} + PDrop\texttt{\scriptsize{(CVPR25)}} & 56.4 & 67.5 & 67.2 & 1945 & 84.2 & 82.0 & 73.7 & 89.9\%  \\
        \hspace{0.5em} + HiRED \texttt{\scriptsize{(AAAI25)}}  & 58.3 & 70.5 & 70.7 & 2086 & 85.4 & 79.6 & 65.4 & 90.8\% \\
        \hspace{0.5em} + SparseVLM\texttt{\scriptsize{(ICML25)}}  & 54.9 & 73.2 & 73.6 & 2047 & 81.1 & 83.2 & 72.2 & 91.9\% \\
        \rowcolor{cyan!7}
        \hspace{0.5em} + \ours\texttt{\scriptsize{(ours)}} & \textcolor{MidnightBlue}{\textbf{59.8}} & \textcolor{MidnightBlue}{\textbf{74.1}} & \textcolor{MidnightBlue}{\textbf{74.2}} & \textcolor{MidnightBlue}{\textbf{2209}} & \textcolor{MidnightBlue}{\textbf{85.5}} & \textcolor{MidnightBlue}{\textbf{83.5}} & \textcolor{MidnightBlue}{\textbf{76.6}} & \textcolor{MidnightBlue}{\textbf{95.8\%}}  \\
        \hline

        \rowcolor{mygray}
        Qwen2-VL-7B & \multicolumn{8}{c}{\textit{Token Reduction} \ $\fg{(\downarrow 88.9\%)}$}\\
        \hspace{0.5em} + FastV \texttt{\scriptsize{(ECCV24)}} & 50.6 & 61.8 & 62.5 & 1760 & 72.2 & 80.5 & 68.6 & 82.2\%  \\
        \hspace{0.5em} + PDrop\texttt{\scriptsize{(CVPR25)}} & 52.5 & 59.2 & 58.3 & 1756 & 74.1 & 80.2 & 67.7 & 81.5\%  \\
        \hspace{0.5em} + HiRED \texttt{\scriptsize{(AAAI25)}}  & 55.5 & 64.3 & 64.3 & 1834 & \textcolor{MidnightBlue}{\textbf{82.8}} & 75.7 & 56.5 & 83.6\% \\
        \hspace{0.5em} + SparseVLM\texttt{\scriptsize{(ICML25)}}  & 47.5 & 59.5 & 61.1 & 1624 & 68.0 & 80.7 & 61.7 & 78.4\% \\
        \rowcolor{cyan!7}
        \hspace{0.5em} + \ours\texttt{\scriptsize{(ours)}} & \textcolor{MidnightBlue}{\textbf{56.1}} & \textcolor{MidnightBlue}{\textbf{69.3}} & \textcolor{MidnightBlue}{\textbf{69.2}} & \textcolor{MidnightBlue}{\textbf{2008}} & 81.2 & \textcolor{MidnightBlue}{\textbf{82.6}} & \textcolor{MidnightBlue}{\textbf{73.4}} & \textcolor{MidnightBlue}{\textbf{90.5\%}} \\
  
        \bottomrule[1.5pt]
	\end{tabular}%
    }
    \label{tab:qwen2vl}
\end{table*}

\noindent \textbf{Generalization to Qwen2-VL-7B.}
Table~\ref{tab:qwen2vl} shows \ours can generalize well to Qwen2-VL-7B under three token ratios. 
\ours achieves 97.6\%, 95.8\% and 90.5\% average accuracy relative to the vanilla model on three ratios. 
Under 66.7\% pruning, \ours achieves the highest average performance (97.6\%), outperforming FastV and PDrop across most benchmarks, with strong results on GQA (61.0), MMB (76.5), POPE (86.7), and SQA (84.2). As compression increases to 77.8\%, \ours maintains the same high average score (95.8\%), demonstrating stable performance despite further token reduction. Even under extreme pruning (88.9\%), \ours continues to outperform competing methods with an average of 90.5\%, showing clear advantages on reasoning-heavy tasks such as GQA, POPE, and VQA$_{\text{Text}}$. 
Across all settings, \ours consistently delivers the best overall performance and exhibits strong robustness as token budgets shrink, validating that cross-layer spectral evolution generalizes beyond LLaVA and remains effective across different MLLMs architectures.

\input{table/image_understanding_llavanext.tex}

\noindent \textbf{Generalization to LLaVA-Next-7B.}
To further evaluate the generalization ability of our method, we conduct comparative experiments on LLaVA-Next-7B. The results are summarized in Table~\ref{tab:main_1_6}. The vanilla model processes 2880 visual tokens and serves as the upper-bound performance reference. Under a strict token budget of 320 tokens (88.9\% token reduction), we compare \ours with several recent token reduction methods including FastV, HiRED, PruMerge, SparseVLM, and PDrop.
Overall, \ours achieves the best performance among all competing methods, reaching 94.7\% of the vanilla performance, significantly outperforming other pruning approaches. In particular, \ours preserves strong performance on reasoning-intensive benchmarks such as GQA (62.6) and MMB (67.1), and maintains competitive results across multimodal evaluation suites including MMB-CN, MME, POPE, and SQA. Compared with other pruning methods that suffer noticeable degradation under aggressive token reduction, \ours consistently maintains higher accuracy across most benchmarks.
Notably, \ours achieves these improvements while operating under the same token budget as other methods, demonstrating that cross-layer spectral evolution provides a more reliable criterion for identifying informative visual tokens. This result indicates that modeling representation dynamics across layers, rather than relying on instantaneous signals, leads to more effective token selection in MLLMs.

\begin{table}[!t]
\caption{{Performance comparison on video understanding benchmarks with Video-LLaVA-7B.}
We consistently prune video tokens from 2048 to 194 (over 90\% reduction). \ours-M denotes \ours with token merging, where pruned tokens are recalled and clustered into compact representatives based on their spectral evolution scores. GPT-4o-mini (GPT) and Gemini-2.5-Flash (Gem) are adopted as assistant evaluators.}
\centering
\begin{adjustbox}{width=1.\columnwidth,center}
\setlength{\tabcolsep}{3pt}
\begin{tabular}{@{}lc cccc cccc cc@{}}
\toprule[1.5pt]
\multirow{2}{*}{\textbf{Method}} & \multirow{2}{*}{\textbf{Eval}}
  & \multicolumn{2}{c}{\textbf{TGIF}}
  & \multicolumn{2}{c}{\textbf{MSVD}}
  & \multicolumn{2}{c}{\textbf{MSRVTT}}
  & \multicolumn{2}{c}{\textbf{ActivityNet}}
  & \multicolumn{2}{c}{\textbf{Avg.}} \\
\cmidrule(lr){3-4}\cmidrule(lr){5-6}\cmidrule(lr){7-8}\cmidrule(lr){9-10}\cmidrule(lr){11-12}
 & & Acc. & Score & Acc. & Score & Acc. & Score & Acc. & Score & Acc. & Score \\
\midrule
\multirow{2}{*}{Video-LLaVA-7B}
  & {\small\textcolor{gray}{GPT}} & 11.3 & 1.19 & 58.9 & 3.31 & 42.7 & 2.69 & 42.0 & 2.20 & 38.7 & 2.35 \\
  & {\small\textcolor{gray}{Gem}} & 14.2 & 0.77 & 57.3 & 2.93 & 47.5 & 2.32 & 42.5 & 2.12 & 40.4 & 2.04 \\
\midrule
\rowcolor{gray!7}\multicolumn{12}{c}{\textit{\footnotesize Hard Pruning}} \\[-2pt]
\midrule

\multirow{2}{*}{FastV}
  & {\small\textcolor{gray}{GPT}} & 2.8 & 1.16 & 35.9 & 2.36 & 31.4 & 2.19 & 34.2 & 1.83 & 26.1 & 1.88 \\
  & {\small\textcolor{gray}{Gem}} & 2.3 & 0.14 & 35.5 & 1.88 & 34.3 & 1.69 & 34.2 & 1.74 & 26.6 & 1.36 \\
\midrule

\multirow{2}{*}{PDrop}
  & {\small\textcolor{gray}{GPT}} & 5.0 & 0.96 & 42.2 & 2.59 & 27.9 & 1.95 & 34.3 & 1.78 & 27.3 & 1.82 \\
  & {\small\textcolor{gray}{Gem}} & 6.1 & 0.37 & 52.2 & 2.54 & 34.6 & 1.65 & 39.4 & 1.93 & 33.0 & 1.62 \\
\midrule

\multirow{2}{*}{FiCoCo-V}
  & {\small\textcolor{gray}{GPT}} & 5.6 & 1.18 & 42.1 & 2.68 & 37.6 & 2.47 & \textcolor{MidnightBlue}{\textbf{43.1}} & 2.23 & 31.9 & 2.13 \\
  & {\small\textcolor{gray}{Gem}} & 6.7 & 0.38 & 51.2 & 2.51 & 44.3 & 2.10 & \textcolor{MidnightBlue}{\textbf{44.9}} & 2.20 & 36.7 & 1.79 \\
\midrule

\rowcolor{cyan!7}
\ours & {\small\textcolor{gray}{GPT}} & \textcolor{MidnightBlue}{\textbf{9.6}} & 1.06 & \textcolor{MidnightBlue}{\textbf{54.1}} & 3.11 & \textcolor{MidnightBlue}{\textbf{39.3}} & 2.56 & 39.9 & 2.03 & \textcolor{MidnightBlue}{\textbf{35.7}} & 2.19 \\
\rowcolor{cyan!7}
  & {\small\textcolor{gray}{Gem}} & \textcolor{MidnightBlue}{\textbf{12.9}} & 0.69 & \textcolor{MidnightBlue}{\textbf{56.3}} & 2.88 & \textcolor{MidnightBlue}{\textbf{47.7}} & 2.27 & 39.9 & 1.99 & \textcolor{MidnightBlue}{\textbf{39.2}} & 1.95 \\
\midrule
\rowcolor{gray!7}\multicolumn{12}{c}{\textit{\footnotesize w/ Token Merging}} \\[-2pt]
\midrule

\multirow{2}{*}{SparseVLM}
  & {\small\textcolor{gray}{GPT}} & 10.7 & 1.03 & 56.0 & 3.20 & 41.9 & 2.63 & 43.6 & 2.21 & 38.1 & 2.27 \\
  & {\small\textcolor{gray}{Gem}} & 13.1 & 0.71 & 58.2 & 3.05 & 48.0 & 2.38 & 44.0 & 2.20 & 40.8 & 2.09 \\
\midrule
\rowcolor{cyan!7}
\ours-M & {\small\textcolor{gray}{GPT}} & \textcolor{MidnightBlue}{\textbf{11.2}} & 1.05 & \textcolor{MidnightBlue}{\textbf{57.0}} & 3.25 & \textcolor{MidnightBlue}{\textbf{42.2}} & 2.61 & \textcolor{MidnightBlue}{\textbf{44.6}} & 2.25 & \textcolor{MidnightBlue}{\textbf{38.7}} & 2.29 \\
\rowcolor{cyan!7}
  & {\small\textcolor{gray}{Gem}} & \textcolor{MidnightBlue}{\textbf{14.7}} & 0.79 & \textcolor{MidnightBlue}{\textbf{60.2}} & 3.07 & \textcolor{MidnightBlue}{\textbf{49.5}} & 2.35 & \textcolor{MidnightBlue}{\textbf{45.0}} & 2.24 & \textcolor{MidnightBlue}{\textbf{42.3}} & 2.11 \\
\bottomrule[1.5pt]
\end{tabular}
\end{adjustbox}
\vspace{-10pt}
\label{tab:main_table_video}
\end{table}

\subsection{Video Understanding Tasks}
\noindent\textbf{Experimental Settings.}
We apply \ours to Video-LLaVA~\cite{lin2024video} and evaluate its performance on four representative video question-answering benchmarks: MSVD-QA~\cite{xu2017video}, MSRVTT-QA~\cite{xu2017video}, TGIF-QA~\cite{jang2017tgif}, and ActivityNet-QA~\cite{yu2019activitynet}. We compare against several advanced baselines: FastV~\cite{chen2024image}, which performs hard pruning; PDrop~\cite{xing2024pyramiddrop}, which adopts a pyramid-shaped progressive pruning schedule; SparseVLM~\cite{zhangsparsevlm} and FiCoCo-V~\cite{FiCoCo2025}, which execute token merging. We report Accuracy (\%) and Score (0--5) using GPT-4o-mini and Gemini-2.5-Flash as assistant evaluators. 

\noindent\textbf{Main Results.}
Table~\ref{tab:main_table_video} reports results when compressing video tokens from 2048 to 194 (over 90\% reduction).
\ours applies per-frame 2D FFT to compute spectral evolution scores and performs hard pruning, as in FastV.
Despite this simplicity, spectral evolution provides a substantially better importance signal: \ours achieves 35.7\% average accuracy, compared to 26.1\% for FastV and 27.3\% for PDrop, demonstrating that per-frame spatial frequency dynamics capture informative visual content more reliably than attention-based selection.
At such extreme compression ratios, hard pruning inevitably discards temporal content that is difficult to recover.
We therefore introduce \ours-M, which recalls the top-scored pruned tokens by their spectral evolution scores and compresses them into compact representatives via density-peak clustering.
\ours-M achieves 38.7\% average accuracy (GPT), matching the vanilla model (38.7\%), and outperforms SparseVLM—which adopts the same token recycling strategy but relies on attention-based scoring—by 0.6 points in accuracy (38.7\% vs.\ 38.1\%), confirming that spectral evolution provides a more reliable signal for identifying critical visual dynamics.

\begin{table*}[t]
\centering
    \caption{
    Efficiency–performance trade-off comparison on LLaVA-1.5-7B and Video-LLaVA-7B under comparable token budgets.
    }
    \scalebox{0.88}{%
    \begin{tabular}{@{}lcccccccc}
       \toprule[1.2pt]
         \multirow{2}{*}{\textbf{Methods}} & \multirow{2}{*}{\textbf{Tokens $\downarrow$}} & \textbf{Prefilling Time $\downarrow$} & \multirow{2}{*}{\textbf{FLOPs $\downarrow$}} & \textbf{KV Cache $\downarrow$}  & \multirow{2}{*}{\textbf{Performance $\uparrow$}} & \textbf{Throughput$\uparrow$} & \textbf{Speedup $\uparrow$} \\
           && \textbf{(MS)} & &\textbf{(MB)} & & \textbf{(samples/s)} & \textbf{(Prefilling)} \\
         \midrule
         \textcolor{gray}{Vanilla LLaVA-1.5-7B} & \textcolor{gray}{576} & \textcolor{gray}{41.9} & \textcolor{gray}{100\%} & \textcolor{gray}{313.0} &\textcolor{gray}{100\%} & \textcolor{gray}{19.4} & \textcolor{gray}{1.00$\times$} \\
         \hspace{0.5em} + FastV & 192 & \textcolor{MidnightBlue}{\textbf{24.8}} & 46.9\% & 121.0 & 92.1\% & \textcolor{MidnightBlue}{\textbf{30.3}} & \textcolor{MidnightBlue}{\textbf{1.69}$\times$} \\
         \hspace{0.5em} + PDrop & 192 & 31.9 & 74.0\% & 211.2 &  96.9\% & 24.6 & 1.31$\times$ \\
         \hspace{0.5em} + SparseVLM & 192 & 33.0 & 49.7\% & 128.2 & 96.3\% & 23.7 & 1.27$\times$ \\

        \rowcolor{cyan!10}
         \hspace{0.5em} + \textbf{\ours}(ours) & 192 & 27.8 & \textcolor{MidnightBlue}{\textbf{43.9}}\% & \textcolor{MidnightBlue}{\textbf{56.0}} & \textcolor{MidnightBlue}{\textbf{99.4\%}} & 27.9 & 1.50$\times$ \\
    \hline

     \textcolor{gray}{Vanilla Video-LLaVA-7B} & \textcolor{gray}{2048} & \textcolor{gray}{114.0} & \textcolor{gray}{100\%} & \textcolor{gray}{1053} &\textcolor{gray}{100\%} & \textcolor{gray}{7.6} & \textcolor{gray}{1.00$\times$} \\
    \hspace{0.5em} + FastV & 194 & \textcolor{MidnightBlue}{\textbf{34.3}} & 20.6\% & 212.9 & 67.4\% & \textcolor{MidnightBlue}{\textbf{19.4}} & \textcolor{MidnightBlue}{\textbf{3.32}$\times$} \\
    \hspace{0.5em} + PDrop & 194 & 45.9 &  20.6\% & 209.2 & 70.5\% & 16.0 & 2.48$\times$ \\
    \hspace{0.5em} + SparseVLM & 194 & 40.0
    & \textcolor{MidnightBlue}{\textbf{10.7}\%} & \textcolor{MidnightBlue}{\textbf{121.1}} & 98.4\% & 18.1 & 2.85$\times$ \\
    \rowcolor{cyan!10}
   \hspace{0.5em} + \textbf{\ours-M}(ours) & 194 & 41.7 & \textcolor{MidnightBlue}{\textbf{10.7}\%} & \textcolor{MidnightBlue}{\textbf{121.1}} & \textcolor{MidnightBlue}{\textbf{100\%}} & 17.3 & 2.73$\times$ \\
    \bottomrule[1.2pt]
    \end{tabular}}
    \label{tab:efficiency}
\end{table*}


\subsection{Analysis}
\noindent \textbf{Efficiency Analysis.}
Table~\ref{tab:efficiency} compares inference efficiency on LLaVA-1.5-7B (192 tokens) and Video-LLaVA-7B (194 tokens).
On images, \ours achieves the lowest FLOPs (\textbf{43.9\%}) and smallest KV cache (\textbf{56.0~MB}) with the highest performance (\textbf{99.6\%}) at 1.50$\times$ speedup; FastV is faster (1.69$\times$) but drops to 92.1\%.
On video, \ours-M matches SparseVLM in FLOPs (\textbf{10.7\%}) and KV cache (\textbf{121.1~MB}) while achieving perfect performance retention (\textbf{100\%} vs.\ 98.4\%) at 2.73$\times$ speedup; FastV again leads in raw speed (3.32$\times$) but suffers severe performance drop.


\vspace{1em}

\begin{figure}[t]
  \centering
  \includegraphics[width=0.99\linewidth]{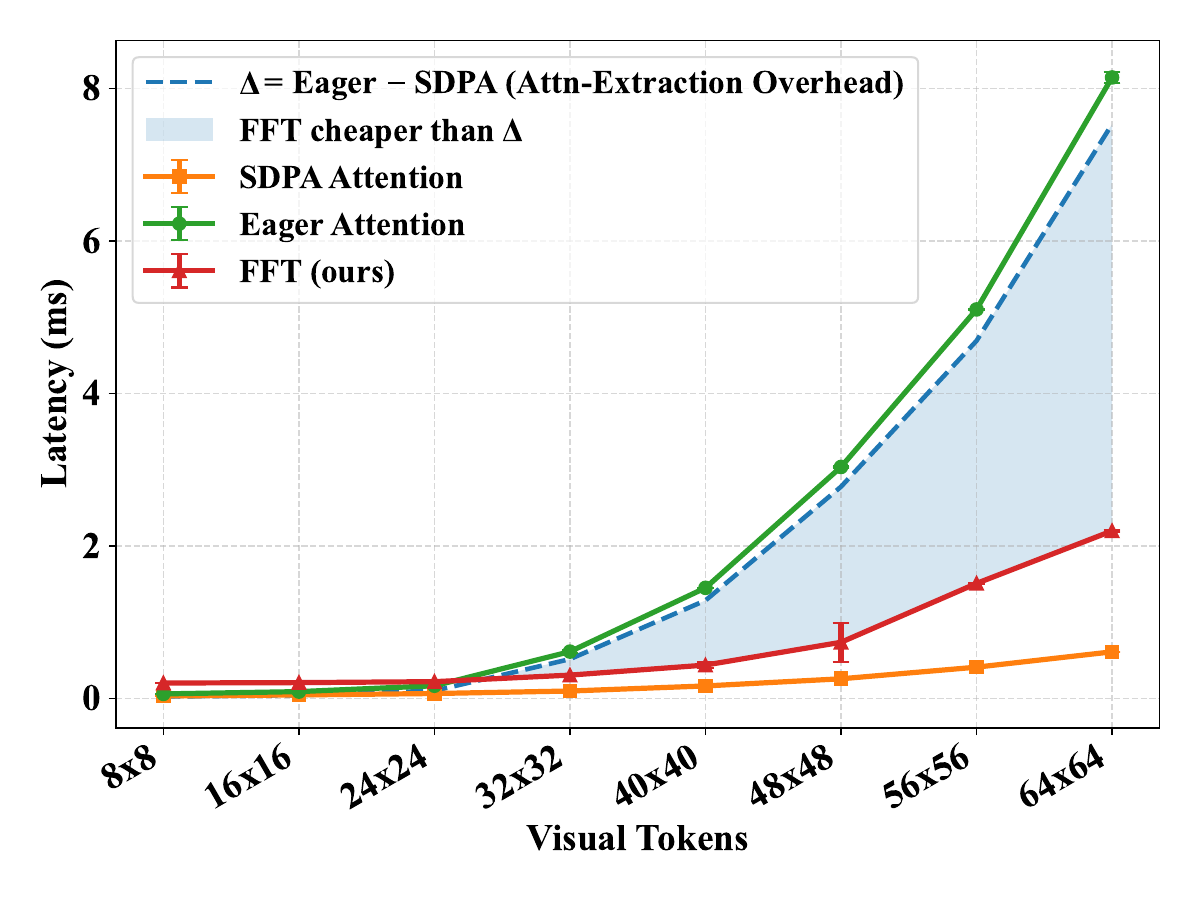}
  \caption{\centering{Latency Comparison to Attention Scoring.}}
  \label{fig:latency-comparison}
\end{figure}

\noindent \textbf{Efficiency Compared with Attention Scoring.} In Fig. \ref{fig:latency-comparison}, we analyze the computational efficiency of our FFT-based scoring relative to the attention-extraction overhead $\Delta$, benchmarked on a single simulated attention layer using LLaMA-3-8B's configuration~\cite{grattafiori2024llama}.
In modern inference engines, optimized kernels such as Scaled Dot-Product Attention (SDPA) achieve high throughput by avoiding the materialization of the full $N \times N$ attention matrix. Consequently, any pruning strategy relying on dense attention weights necessitates falling back to a sub-optimal ``Eager'' implementation, which reintroduces a quadratic latency penalty $O(N^2)$. As illustrated, while the extraction overhead $\Delta$ (defined as $\text{Latency}_{\text{Eager}} - \text{Latency}_{\text{SDPA}}$) escalates quadratically with sequence length $N$, our \ours module maintains a near-linear $O(N \log N)$ complexity due to the efficiency of the Fast Fourier Transform. Even at a spatial resolution of $64 \times 64$ ($N=4096$), the latency of our spectral scoring remains significantly below the $\Delta$ threshold. This gap demonstrates that spectral evolution also offers a more hardware-efficient pathway for token reduction.

\begin{figure*}[t]
  \centering
  \begin{subfigure}[b]{0.33\linewidth}
    \centering
    \includegraphics[width=\linewidth]{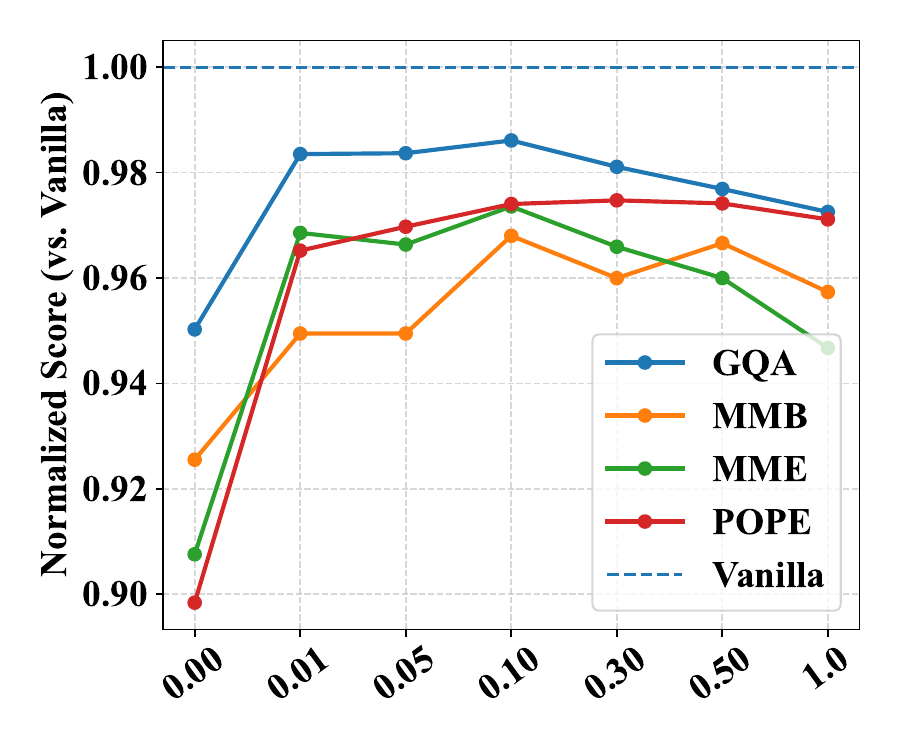}
    \caption{Cutoff Ratio $r$}
    \label{fig:ablation-a}
  \end{subfigure}\hfill
  \begin{subfigure}[b]{0.33\linewidth}
    \centering
    \includegraphics[width=\linewidth]{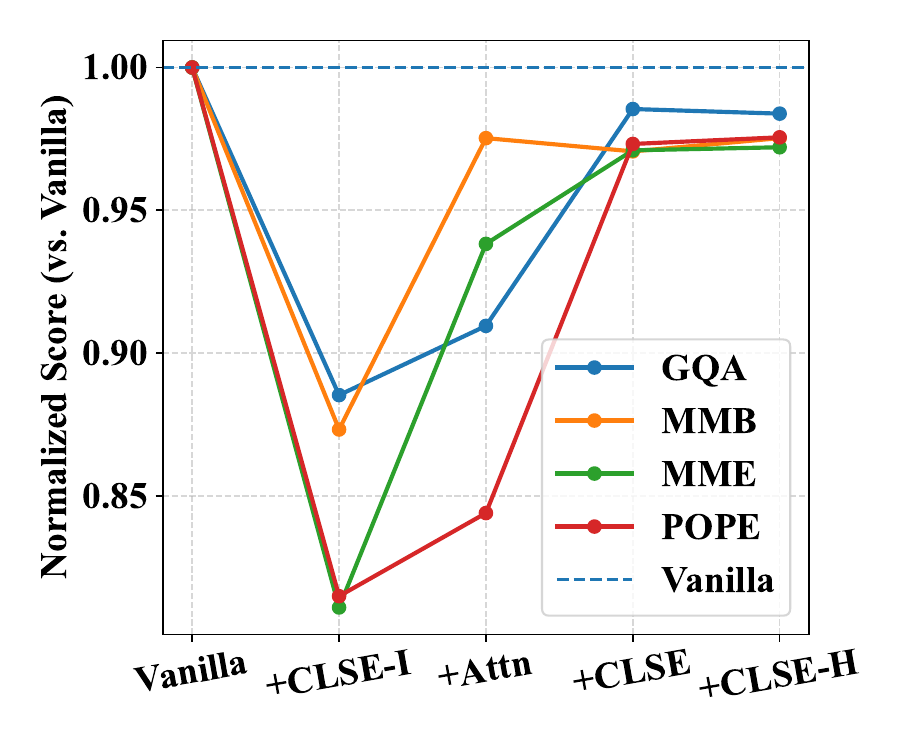}
    \caption{Spectral Evolution} 
    \label{fig:ablation-b}
  \end{subfigure}\hfill
  \begin{subfigure}[b]{0.33\linewidth}
    \centering
    \includegraphics[width=\linewidth]{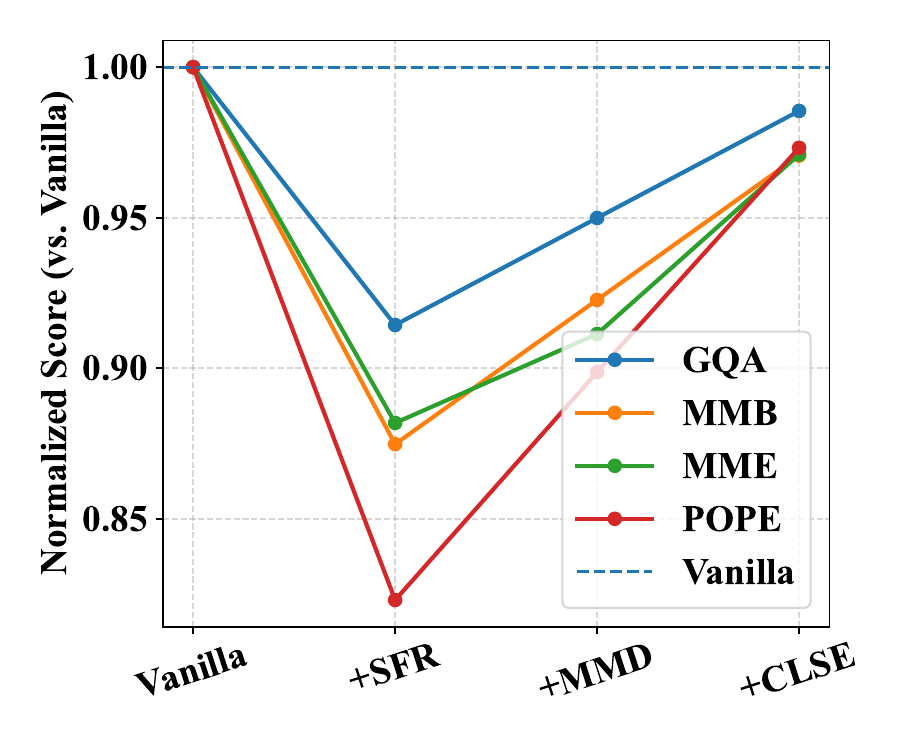}
    \caption{Token Importance Metrics}
    \label{fig:ablation-c}
  \end{subfigure}
  \caption{Ablation studies on LLaVA-1.5-7B pruned after the first layer of LLM decoder with 192 retained tokens. Scores are normalized to the Vanilla model. (a) Effect of the Gaussian high-pass cutoff ratio. (b) Effect of spectral evolution and its integration with attention. (c) Impact of the magnitude measures for cross-layer component analysis. }
  \label{fig:ablation}

\end{figure*}

\vspace{1em}
\noindent \textbf{Ablation Studies.}
Fig.~\ref{fig:ablation} presents comprehensive ablation analyses 
on LLaVA-1.5-7B with 192 retained tokens. All results are normalized to the full-token model, where 1.0 indicates matching full performance.
Fig.~\ref{fig:ablation}(a) evaluates the impact of the Gaussian high-pass cutoff parameter. In all experimental scenarios, we set the cutoff to 0.1. Performance improves substantially when moving from $r=0.00$ to small values, indicating that mild suppression of low-frequency components effectively reduces redundancy. Across benchmarks, moderate cutoff values consistently yield the strong \ours results.
Fig.~\ref{fig:ablation}(b) evaluates the core components of our framework. We observe a catastrophic performance collapse in \ours-Inertia (\ours-I), which retains tokens with the \textit{lowest} spectral evolution scores (the most static tokens), serving as counterfactual evidence that depth-wise spectral evolution is the primary driver of semantic activity.
In contrast, while pure attention-based pruning suffers from positional bias, our \ours criterion significantly stabilizes performance. To further fortify the system, \ours-Hybrid (\ours-H) integrates \ours with attention weights that can produce on-par performance like \ours.
Fig.~\ref{fig:ablation}(c) evaluates \ours against alternative depth-wise metrics.
We compare with Spatial Feature Residual (SFR), which measures the L1-norm of feature residuals, and Mean Magnitude Difference (MMD), which monitors changes in channel-wise mean activations.
SFR exhibits performance collapse as its reliance on raw pixel-level intensity makes it susceptible to high-frequency artifacts.
MMD, while slightly more stable, remains a coarse-grained heuristic. By collapsing high-dimensional token dynamics into a scalar magnitude, it fails to capture the structural redistribution occurring within the frequency spectrum.
In contrast, \ours consistently establishes a new performance upper bound.

\begin{figure*}[t]
    \centering
    \includegraphics[width=1\linewidth]{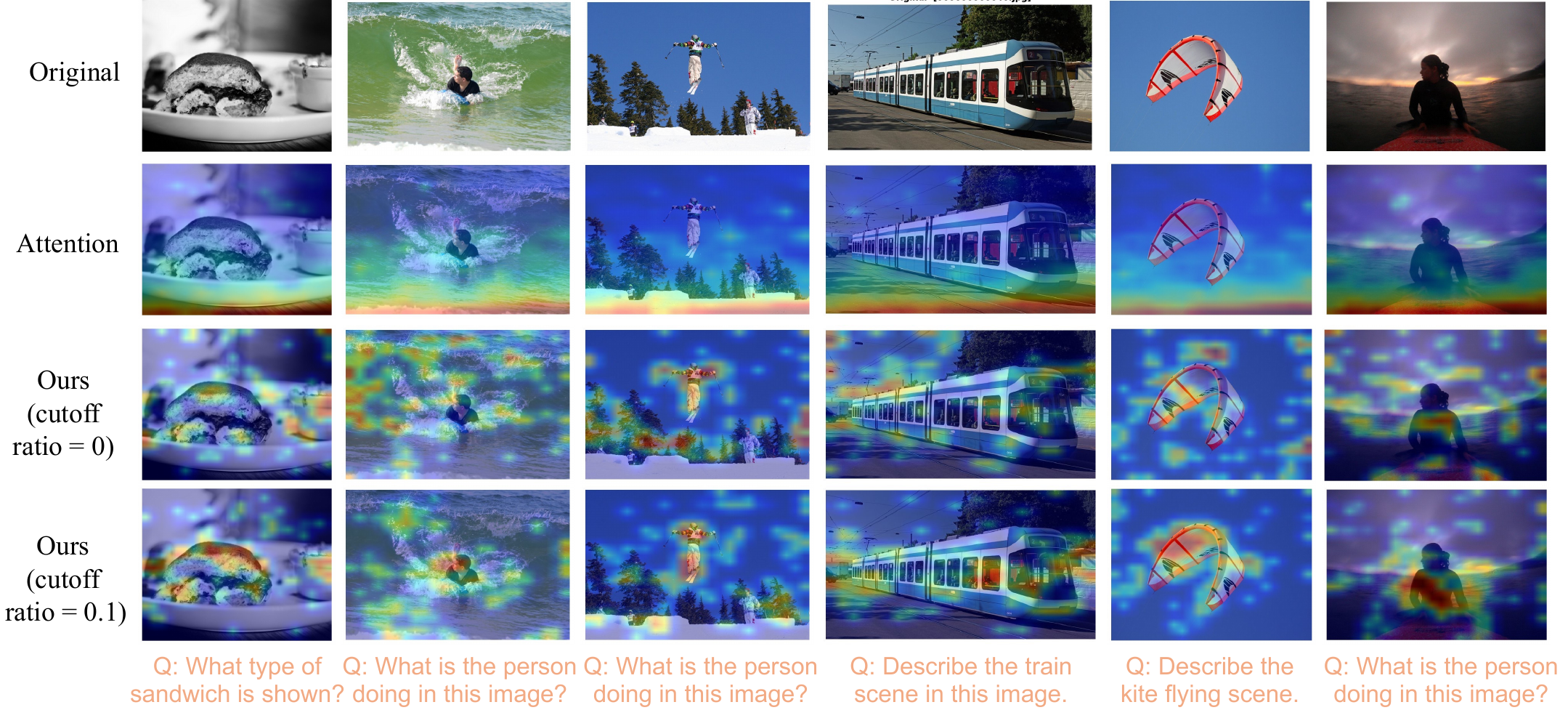}
    \vspace{-10pt}
    \caption{Qualitative comparison of token importance maps. Top row: original images. 
Second row: attention-based token scores. Third and fourth rows: \ours without frequency suppression ($r=0$) and with moderate suppression ($r=0.1$), respectively. 
Compared to attention, \ours highlights structurally evolving and semantically relevant regions while suppressing redundant background areas. Moderate frequency filtering further improves focus on task-relevant objects.
}
    \label{fig:sample}
    \vspace{-10pt}
\end{figure*}

\vspace{0.5em}
\noindent \textbf{Qualitative Analysis.}
Fig.~\ref{fig:sample} visualizes token importance maps under different strategies. 
Attention-based scores tend to spread over large background regions and occasionally over-emphasize visually salient but semantically irrelevant areas. 
In contrast, \ours concentrates on structurally meaningful regions that actively evolve across layers, such as key objects or human subjects. 
When the cutoff ratio is set to $r=0.1$, the maps exhibit sharper localization and reduced background noise compared to $r=0$, confirming that moderate suppression of low-frequency components enhances semantic discrimination. 
These visualizations support our quantitative findings that cross-layer spectral evolution provides a more reliable token importance signal than raw attention.

\vspace{0.5em}
\noindent \textbf{Additional Experiments in Appendix.}
We provide complementary analyses in the appendix to further validate \ours. 
These include experimental setup details (\textit{e.g.,} baseline methods, implementation, and datasets), 
details of a \ours variant with token merging, 
ablation on pruning at different layers, 
experiments on additional MLLMs (InternVL, Qwen2-VL-72B), 
complete numeric results for LLaVA-1.5-13B, 
a detailed computational complexity analysis,
channel dimension impact on run time, 
and impact of 3D vs 2D FFT on video MLLMs.

\section{Conclusion}
In this work, we introduced a token pruning framework, namely Cross-Layer Spectral Evolution (\ours), that models how token representations evolve across depth, capturing the structural transition from high-frequency detail to low-frequency semantic abstraction. 
Extensive experiments on image and video benchmarks demonstrate that \ours consistently achieves superior accuracy–efficiency trade-offs across multiple architectures and compression levels. 
In particular, the performance gap widens under aggressive pruning, highlighting the robustness of spectral evolution as a token importance signal. 
We hope this work encourages further exploration of spectral representations for efficient multimodal reasoning.

\section*{Acknowledgements}
This work is supported by the National Natural Science Foundation of China under Grant No. 62502429, and the Zhejiang Key Laboratory Project (2024E10001) 



%
%
\bibliographystyle{splncs04}
\bibliography{main}

\section*{Appendix}

\appendixtoc

\begingroup
\let\origaddcontentsline\addcontentsline
\renewcommand{\addcontentsline}[3]{%
  \def\tempa{#1}\def\tempb{toc}%
  \ifx\tempa\tempb
    \origaddcontentsline{apx}{#2}{#3}
  \else
    \origaddcontentsline{#1}{#2}{#3}
  \fi
}
\setcounter{section}{0}
\renewcommand{\thesection}{\Alph{section}}

\setcounter{subsection}{0}
\renewcommand{\thesubsection}{\thesection.\arabic{subsection}}
\section{Implementation Details}

\subsection{Datasets}
\label{app:dataset}

Our experiments are conducted on a suite of widely recognized benchmarks spanning image understanding and video understanding, each targeting a distinct aspect of multimodal intelligence.

\vspace{0.5em}
\noindent\textit{\textbf{Image Understanding Benchmarks.}}
\vspace{0.3em}

\noindent\textbf{GQA}~\cite{hudson2019gqa} targets visual scene understanding through a structured combination of scene graphs, image content, and compositional questions. Each image is accompanied by detailed spatial annotations and per-object attribute information, enabling fine-grained grounding of question-answer pairs. The benchmark is designed to probe a model's reasoning over object relationships, spatial layouts, and scene-level semantics beyond simple recognition.

\noindent\textbf{MMBench}~\cite{liu2024mmbench} adopts a three-tier capability taxonomy to systematically profile multimodal models. At the broadest level, it distinguishes perception from reasoning; the intermediate level refines these into six cognitive sub-abilities; and the most granular level further decomposes performance across 20 fine-grained dimensions. This structured hierarchy enables interpretable and multi-faceted analysis of model strengths and weaknesses. MMBench-CN is a Chinese-language variant built upon the same evaluation framework.

\noindent\textbf{MME}~\cite{fu2025mme} offers a systematic diagnostic of perceptual and cognitive abilities through 14 targeted subtasks, each defined by carefully formulated instruction-answer pairs. The benchmark prioritizes brevity and precision in its instructions to mitigate data leakage risks and ensure evaluation fairness. Spanning both low-level visual perception and high-level reasoning, MME provides a thorough and reliable measure of multimodal model capabilities.

\noindent\textbf{POPE}~\cite{li2023evaluating} addresses the problem of object hallucination through a binary probing paradigm: models are presented with yes/no questions about the existence of specific objects in an image. Performance is measured by accuracy, recall, precision, and F1 score across three sampling strategies (random, popular, and adversarial), enabling a controlled and multi-faceted analysis of hallucination tendencies in multimodal systems.

\noindent\textbf{ScienceQA}~\cite{lu2022learn} is a large-scale multimodal benchmark drawing from elementary and middle school science curricula across three major domains: natural science, social science, and language science. Questions are organized hierarchically into 26 topics, 127 categories, and 379 fine-grained skills, yielding a richly structured evaluation space. Each question is accompanied by a natural language explanation and, for a subset, relevant context images, enabling assessment of not only answer accuracy but also multi-step reasoning chains and cross-modal interpretation. The breadth of subject matter combined with the depth of the skill taxonomy makes it a demanding testbed for cross-domain generalization.

\noindent\textbf{TextVQA}~\cite{singh2019towards} probes a model's capacity to jointly reason over visual scenes and the text embedded within them. Unlike purely visual QA tasks, answering TextVQA questions demands that a model first localize and read text regions in the image, then integrate that textual content with the broader visual context to produce accurate responses. This dual dependency on visual perception and text recognition makes it a particularly challenging benchmark at the intersection of OCR and visual reasoning.

\noindent\textbf{VizWiz}~\cite{gurari2018vizwiz} is grounded in a real-world accessibility scenario: images were taken by blind individuals using smartphone cameras, and the accompanying questions reflect their actual information needs. The dataset comprises 20,523 training, 4,319 validation, and 8,000 test image-question pairs, each annotated by 10 crowd-sourced workers. A distinctive challenge is that a non-trivial fraction of questions are inherently unanswerable due to image quality issues, requiring models to simultaneously assess question answerability and produce accurate responses when possible.

\noindent\textbf{OCRBench}~\cite{liu2024ocrbench} provides a systematic evaluation of text perception and understanding across five task categories: text recognition in natural scenes, scene text-centric VQA, document-oriented VQA, key information extraction, and handwritten mathematical expression recognition. By spanning this diverse range of text-centric challenges, it enables thorough assessment of a multimodal model's proficiency in handling textual content encountered across real-world visual environments.

\begin{figure*}[!htbp]
\centering
\setlength{\tabcolsep}{6pt}
\renewcommand{\arraystretch}{1.0}
\newlength{\qualht}\setlength{\qualht}{7.2cm}
\begin{tabular}{ccc}

\begin{minipage}[t][{\qualht}][t]{0.30\textwidth}\centering
  {\small\textbf{GQA}}\\[2pt]
  \includegraphics[width=\linewidth,height=\linewidth,keepaspectratio]{figures/qualitative/gqa}\\[3pt]
  {\fontsize{7.5}{9.5}\selectfont\raggedright\textbf{Q:} Are there beds next to the small outlet?\\\textbf{A:} \textcolor{ForestGreen}{yes}}
\end{minipage}
&
\begin{minipage}[t][{\qualht}][t]{0.30\textwidth}\centering
  {\small\textbf{MMBench}}\\[2pt]
  \includegraphics[width=\linewidth,height=\linewidth,keepaspectratio]{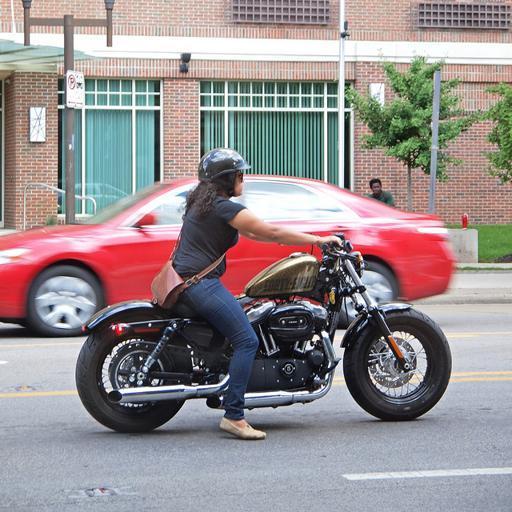}\\[3pt]
  {\fontsize{7.5}{9.5}\selectfont\raggedright\textbf{Q:} Which one is the correct caption of this image?\\\textbf{A:} \textcolor{ForestGreen}{A woman is riding a motorcycle down the street.}}
\end{minipage}
&
\begin{minipage}[t][{\qualht}][t]{0.30\textwidth}\centering
  {\small\textbf{MMBench-CN}}\\[2pt]
  \includegraphics[width=\linewidth,height=\linewidth,keepaspectratio]{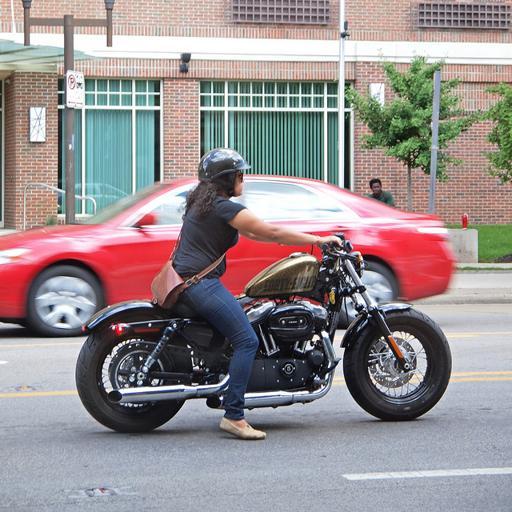}\\[3pt]
  {\fontsize{7.5}{9.5}\selectfont\raggedright\textbf{Q:} \begin{CJK}{UTF8}{gbsn}以下哪个是这张图片的正确标题？\end{CJK}\\\textbf{A:} \textcolor{ForestGreen}{\begin{CJK}{UTF8}{gbsn}一个女人在街上骑摩托车。\end{CJK}}}
\end{minipage}
\\[8pt]

\begin{minipage}[t][{\qualht}][t]{0.30\textwidth}\centering
  {\small\textbf{MME}}\\[2pt]
  \includegraphics[width=\linewidth,height=\linewidth,keepaspectratio]{figures/qualitative/mme}\\[3pt]
  {\fontsize{7.5}{9.5}\selectfont\raggedright\textbf{Q:} Are there two bowls in this image?\\\textbf{A:} \textcolor{ForestGreen}{Yes}}
\end{minipage}
&
\begin{minipage}[t][{\qualht}][t]{0.30\textwidth}\centering
  {\small\textbf{POPE}}\\[2pt]
  \includegraphics[width=\linewidth,height=\linewidth,keepaspectratio]{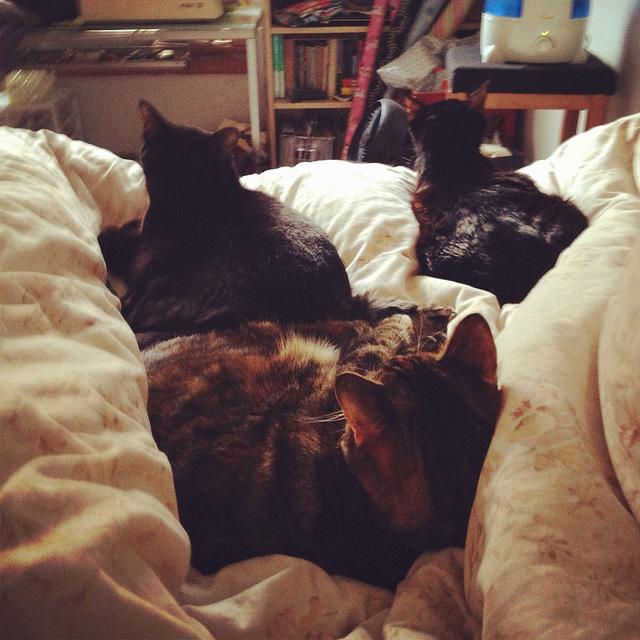}\\[3pt]
  {\fontsize{7.5}{9.5}\selectfont\raggedright\textbf{Q:} Is there a bed in the image?\\\textbf{A:} \textcolor{ForestGreen}{yes}}
\end{minipage}
&
\begin{minipage}[t][{\qualht}][t]{0.30\textwidth}\centering
  {\small\textbf{ScienceQA}}\\[2pt]
  \includegraphics[width=\linewidth,height=\linewidth,keepaspectratio]{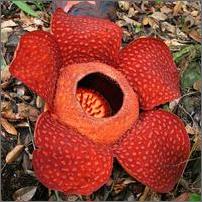}\\[3pt]
  {\fontsize{7.5}{9.5}\selectfont\raggedright\textbf{Q:} Is \textit{Rafflesia arnoldii} made up of many cells?\\\textbf{A:} \textcolor{ForestGreen}{yes}}
\end{minipage}
\\[8pt]

\begin{minipage}[t][{\qualht}][t]{0.30\textwidth}\centering
  {\small\textbf{TextVQA}}\\[2pt]
  \includegraphics[width=\linewidth,height=\linewidth,keepaspectratio]{figures/qualitative/textvqa}\\[3pt]
  {\fontsize{7.5}{9.5}\selectfont\raggedright\textbf{Q:} What kind of beer is this?\\\textbf{A:} \textcolor{ForestGreen}{ale}}
\end{minipage}
&
\begin{minipage}[t][{\qualht}][t]{0.30\textwidth}\centering
  {\small\textbf{VizWiz}}\\[2pt]
  \includegraphics[width=\linewidth,height=\linewidth,keepaspectratio]{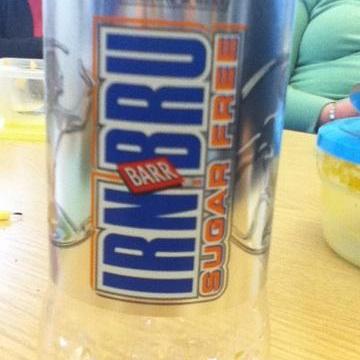}\\[3pt]
  {\fontsize{7.5}{9.5}\selectfont\raggedright\textbf{Q:} What is the name of the drink?\\\textbf{A:} \textcolor{ForestGreen}{irn bru}}
\end{minipage}
&
\begin{minipage}[t][{\qualht}][t]{0.30\textwidth}\centering
  {\small\textbf{OCRBench}}\\[2pt]
  \includegraphics[width=\linewidth,height=\linewidth,keepaspectratio]{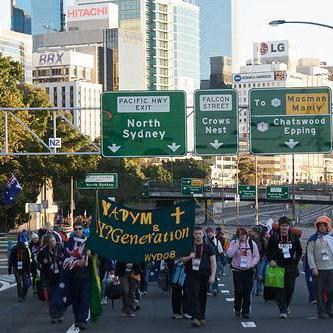}\\[3pt]
  {\fontsize{7.5}{9.5}\selectfont\raggedright\textbf{Q:} What is the Mosman Manly exit going to?\\\textbf{A:} \textcolor{ForestGreen}{Chatswood Epping}}
\end{minipage}

\end{tabular}
\caption{Representative examples from each image understanding benchmark used in our evaluation. Each cell shows the benchmark name, a sample image, the question posed to the model, and the ground-truth answer.}
\label{fig:qualitative_examples}
\end{figure*}

\vspace{0.5em}
\noindent\textit{\textbf{Video Understanding Benchmarks.}}
\vspace{0.3em}

\noindent\textbf{TGIF-QA}~\cite{jang2017tgif} extends visual question answering to the video domain using animated GIFs sourced from Tumblr, comprising 165,000 QA pairs. It defines four task types that collectively demand temporal understanding beyond what static image QA can assess: repetition counting (how many times an action occurs), repeating action identification (what action is repeated a given number of times), state transition reasoning (what changes between two temporal states), and frame-level open-ended QA. This multi-task design makes TGIF-QA a comprehensive benchmark for evaluating both temporal and spatial reasoning in video understanding models.

\noindent\textbf{MSVD-QA}~\cite{chen2011collecting,xu2017video} builds on the Microsoft Research Video Description corpus, transforming its 1,970 short video clips into an open-ended question answering benchmark with approximately 50.5K QA pairs. Questions are organized into five interrogative categories (\textit{what}, \textit{who}, \textit{how}, \textit{when}, and \textit{where}), covering a broad spectrum of video content and temporal aspects, and supporting evaluation across both video QA and video captioning scenarios.

\noindent\textbf{MSRVTT-QA}~\cite{xu2016msr,xu2017video} scales video QA evaluation to 10,000 clips and 243,000 QA pairs drawn from the MSR-VTT dataset. Its central challenge is that accurate responses require models to jointly process visual appearance and temporal dynamics across the clip, neither alone is sufficient. Sharing the same five question-type taxonomy as MSVD-QA (\textit{what}, \textit{who}, \textit{how}, \textit{when}, \textit{where}), it serves as a higher-volume and more diverse complement for benchmarking video comprehension.

\noindent\textbf{ActivityNet-QA}~\cite{yu2019activitynet} is a large-scale benchmark designed for understanding complex web videos through question answering. It contains 58K human-annotated QA pairs over 5,800 video clips sourced from the ActivityNet dataset, covering a wide range of real-world human activities. Questions require temporal reasoning and holistic video comprehension, making it a challenging testbed for long-form video understanding.

\subsection{Baselines}
\label{app:baselines}
We compare against a range of representative token-reduction methods for accelerating multimodal large language models (MLLMs). Despite sharing the common goal of eliminating redundant visual tokens, these methods differ substantially in their strategies, spanning token merging, attention-guided pruning, and adaptive budget allocation.

\noindent\textbf{ToMe}~\cite{bolyatoken} consolidates redundant visual representations directly within transformer attention blocks by applying a bipartite soft-matching procedure at each layer. Tokens are partitioned into two complementary sets, and the most similar pairs across the boundary are selected and fused into unified embeddings, carrying proportionally combined features rather than discarding information outright. By folding this matching step into the standard forward pass, ToMe achieves near-linear throughput improvements as a function of the merge ratio with negligible additional memory overhead. No task-specific retraining or architectural modification is required, making it immediately applicable to any vision transformer backbone. This training-free property and architectural agnosticism make ToMe a general-purpose compression primitive.

\noindent\textbf{FastV}~\cite{chen2024image} targets the computational bottleneck introduced by long visual token sequences by pruning low-value tokens at an early inference stage inside the LLM. Token importance is assessed using attention maps produced by the language model itself, rather than the upstream vision encoder, providing query-conditioned relevance estimates that are directly grounded in how the model processes the current input. Tokens that accumulate little attention across the early LLM layers are judged uninformative for the current query and discarded before reaching computationally heavier deeper layers. Because the pruning decision is made at inference time without any parameter updates, FastV integrates into existing MLLM pipelines as a plug-and-play module. The method exploits the empirical regularity that LLM attention is strongly concentrated on a sparse subset of visual tokens from the earliest layers, yielding substantial FLOPs reduction with modest accuracy degradation.

\noindent\textbf{SparseVLM}~\cite{zhangsparsevlm} formulates token selection as a cross-modal relevance problem, deriving per-token importance scores from the interaction between textual query representations and visual features. A key design choice is the adoption of sample-adaptive sparsity: rather than imposing a fixed retention ratio uniformly across all inputs, the method calibrates pruning aggressiveness to the estimated informational density of each image, retaining more tokens for visually complex or query-ambiguous scenes and pruning more aggressively when semantic content is concentrated. The approach further incorporates a token recycling mechanism to partially compensate for information loss—discarded tokens are not simply dropped but are redistributed into the retained set through a lightweight aggregation step, allowing low-attention regions to contribute to the final representation in compressed form. The combination of cross-modal scoring, content-adaptive budgeting, and partial information recovery enables SparseVLM to sustain strong task performance under significant compression without any model fine-tuning.

\noindent\textbf{HiRED}~\cite{arif2025hired} addresses the token budget problem in high-resolution MLLMs that decompose inputs into a global thumbnail paired with multiple local crops, where naive uniform allocation wastes representational resources on uninformative regions. Its compression strategy proceeds in two hierarchically nested stages: at the partition level, each crop's token budget is set in proportion to its \texttt{CLS} attention score, ensuring that image regions attracting greater query-driven attention receive correspondingly more representation capacity. Within each budget-allocated partition, the locally most informative tokens are then retained and the remainder pruned, preserving fine-grained spatial structure where it is most needed. This two-stage design—global budget distribution followed by local token selection—maintains spatial awareness throughout the reduction process, a property that flat, context-insensitive methods cannot provide. HiRED is consequently well-suited to tasks involving fine-grained localization or structured scene understanding, where the informational value of image sub-regions is highly non-uniform.

\noindent\textbf{LLaVA-PruMerge}~\cite{shang2025llava} integrates pruning and merging into a sequential two-stage compression pipeline tailored to the architectural conventions of LLaVA-style models. In the first stage, token importance is estimated through sparse \texttt{CLS}-to-visual cross-attention: tokens that attract low aggregate attention from the class token are flagged as low-priority candidates and removed from the sequence. In the second stage, the surviving tokens are clustered according to the pairwise similarity of their key vectors, and tokens within each cluster are fused into a single unified embedding, reducing the token count further without discarding the information they collectively encode. This two-phase design achieves compression at two independent granularities—coarse-level elimination through pruning and fine-level consolidation through key-similarity merging—allowing the method to operate effectively under tight token budgets. The entire pipeline is training-free and introduces no additional parameters, relying exclusively on attention-derived and similarity-derived signals that are already available during the forward pass.

\noindent\textbf{PyramidDrop}~\cite{xing2024pyramiddrop} applies a staged compression schedule in which the proportion of tokens dropped increases progressively with layer depth, resulting in a pyramid-shaped token count profile across the transformer stack. The rationale is grounded in the observation that early layers process relatively distributed feature maps where broad coverage is beneficial, while deeper layers operate on increasingly abstracted, task-focused representations where a smaller, more focused token set is sufficient. By retaining a fuller complement of tokens in shallow layers and incrementally discarding a larger fraction at successive depth checkpoints, PyramidDrop avoids the abrupt representational discontinuity characteristic of single-stage pruning schemes. This depth-aware schedule allocates computational resources proportionally to the layer-wise marginal value of token diversity, delivering significant aggregate FLOPs savings across the full network. The pyramid structure thereby achieves a more principled trade-off between efficiency and representational fidelity than uniform or single-point pruning.

\noindent\textbf{FiCoCo}~\cite{han2024rethinking} addresses token redundancy through a unified three-stage \emph{filter\allowbreak--\allowbreak{}correlate\allowbreak--\allowbreak{}compress} pipeline that tackles different sources of excess representation in a principled sequence. The filter stage performs an initial coarse selection, removing tokens whose predicted contribution to downstream task performance falls below a lightweight salience threshold, eliminating the most obvious redundancy at low cost. The correlate stage then analyses pairwise relationships among surviving tokens—both within the visual modality and across text-visual boundaries—to identify clusters of tokens whose encoded information substantially overlaps. The compress stage aggregates each identified cluster into a single compact representative, further condensing the sequence while retaining the essential semantic content of the group. By explicitly decoupling these three operations, FiCoCo can address complementary forms of redundancy independently and in sequence, enabling more thorough compression than single-criterion methods that optimize for only one dimension of token reduction at a time.

\subsection{Experimental Setup}
\label{app:experimental_setup}

All experiments are conducted on a server equipped with six NVIDIA RTX PRO 6000 Blackwell Server Edition GPUs, each with 96\,GB of VRAM, providing ample capacity to host large-scale vision-language models without cross-GPU tensor parallelism for most configurations. The software environment consists of Python 3.10, PyTorch 2.9.1, CUDA 12.8, and cuDNN 9.1.0. All models are loaded and evaluated in \texttt{float16} precision. Inference is performed with a batch size of one to faithfully reflect single-sample latency, which is the practically relevant setting for interactive MLLM deployment. All baselines are reproduced under the same hardware and software stack using their official codebases. No task-specific fine-tuning is applied to any method; all token-reduction decisions are made at inference time.

\subsection{Method Pipeline of CLSE with Token-Merging}
\label{app:method_pipeline}

We describe the complete pipeline of \ours, covering token importance scoring, rank-adaptive pruning, and the merging step that recovers information from discarded tokens.

\noindent\textbf{Overall Flow.}
At a designated pruning layer $\ell$, the model (i) runs a normal forward pass to obtain hidden states before and after the layer, (ii) computes a per-token importance score from the resulting representations, (iii) retains the top-ranked tokens and discards the rest, (iv) optionally merges a subset of discarded tokens into compact representatives, and (v) reassembles the sequence for all subsequent layers.

\noindent\textbf{Per-Frame 2D FFT vs.\ 3D Spatiotemporal FFT.}
A natural question for video inputs is whether a 3D FFT over the full spatiotemporal volume could capture richer temporal structure than per-frame 2D FFT.
We compare both variants on Video-LLaVA-7B~\cite{lin2024video}, where $\text{CLSE}_{2\text{df}}$ applies 2D FFT independently to each sampled frame and $\text{CLSE}_{3\text{d}}$ applies a single 3D FFT across all frames.
As shown in Table~\ref{tab:ablation_2d_vs_3d}, per-frame 2D FFT substantially outperforms its 3D counterpart on every benchmark and evaluator.
This is expected: with only $T\!=\!8$ sparsely sampled frames, the temporal dimension is far too short for reliable 3D spectral analysis, and temporal coherence is more effectively captured by the cross-layer evolution factor.
3D FFT also loses the ability to process frames in parallel, introducing additional computational overhead without accuracy benefit.
\input{table/fft_2d_vs_3d_ablation}

\noindent\textbf{Cross-Layer Spectral Evolution (CLSE).}
The CLSE score measures how much each token's spectral structure changes between adjacent layers; its computation is detailed in the main paper. The final per-token importance multiplies the text-to-visual attention score by the CLSE factor. The number of tokens to retain is determined adaptively by the numerical rank of the text-visual attention matrix $\mathbf{A}_{\text{text}\to\text{vis}}$, rather than a fixed ratio: the top-$r$ tokens by joint score are kept, where $r = \mathrm{rank}(\mathbf{A}_{\text{text}\to\text{vis}})$.

\noindent\textbf{Density-Peak Clustering Merge.}
Pruned tokens are not simply discarded. A two-stage recycling mechanism recovers informative tokens from the pruned set. In \textbf{Stage~1}, we select the top 30\% of pruned tokens by joint score, filtering out those with negligible attention. In \textbf{Stage~2}, we apply density-peak clustering to compress these $K$ recovered tokens into $\lfloor K/10 \rfloor + 1$ representative embeddings. Cluster centers are identified by jointly maximizing local density and distance to the nearest higher-density neighbor; each remaining token is then assigned to its closest center and compressed by uniform averaging within the cluster. The final reassembled sequence takes the form:
\begin{equation}
\bigl[\underbrace{\text{pre-prompt}}_{\text{unchanged}} \mid \underbrace{\text{kept visual}}_{r\text{ tokens}} \mid \underbrace{\text{merged reps}}_{\lfloor K/10\rfloor+1\text{ tokens}} \mid \underbrace{\text{text tokens}}_{\text{unchanged}}\bigr].
\end{equation}
This design preserves discriminative high-frequency tokens at full resolution while representing redundant-but-informative tokens compactly, achieving a favorable balance between compression ratio and information retention.

\section{Computational Efficiency}
\label{app:efficiency}

\subsection{Theoretical FLOPs}
\label{app:flops}

To characterize the computational cost of MLLMs, we analyze the two dominant components: the self-attention mechanism and the feed-forward network (FFN). The total FLOPs of a transformer with $T$ layers, sequence length $n$, hidden dimension $d$, and FFN intermediate size $m$ can be written as:
\begin{equation}
\text{Total FLOPs} = T \times \left(4nd^2 + 2n^2d + 2ndm\right).
\end{equation}
This expression reveals the quadratic dependence on sequence length $n$, motivating token reduction as an effective strategy for inference acceleration. Following FastV~\cite{chen2024image}, we estimate FLOPs under token pruning as:
\begin{equation}
\begin{split}
\text{Post-Pruning FLOPs} = \;&L\times\!\left(4nd^2+2n^2d+2ndm\right) \\
+\;&(T-L)\times\!\left(4\hat{n}d^2+2\hat{n}^2d+2\hat{n}dm\right),
\end{split}
\end{equation}
where $L$ is the pruning layer and $\hat{n}$ is the sequence length after pruning. The theoretical FLOPs reduction ratio is then:
\begin{equation}
1 - \frac{\text{Post-Pruning FLOPs}}{\text{Total FLOPs}}.
\end{equation}

\noindent\textbf{FFT Scoring Overhead.}
The CLSE scoring step introduces an additional per-inference cost from the per-frame 2D FFT. Let $N_v$ denote the total number of visual tokens, $F$ the number of sampled frames, and $N_f = N_v / F$ the spatial tokens per frame. For each frame, a 2D FFT is applied to the $N_f$-point spatial grid independently across all $d$ feature channels. A 2D FFT over $N_f$ points (computed via \texttt{torch.fft.fft2}) requires $\mathcal{O}(N_f\log_2 N_f)$ operations per channel. Summing over all frames and channels yields:
\begin{equation}
\text{FFT FLOPs} = \mathcal{O}\!\left(d\,N_v\,\log_2(N_v / F)\right).
\end{equation}
For image inputs ($F=1$) this reduces to $\mathcal{O}(dN_v\log_2 N_v)$. To contextualize this cost, consider LLaVA-1.5-7B~\cite{liu2024improved} with $d{=}4096$, $N_v{=}576$, and $F{=}1$: the FFT scoring step incurs approximately $108$\,M FLOPs, whereas a single transformer layer costs ${\approx}97$\,B FLOPs under the formula above ($n{=}600$, $m{=}11008$). The FFT overhead thus constitutes less than $0.12\%$ of a single layer and under $0.004\%$ of the total forward-pass cost, confirming that the spectral scoring step is computationally negligible.

\subsection{Latency Measurement}
\label{app:latency}

We measure wall-clock latency using \texttt{torch.cuda.Event} to ensure precise GPU-side timing. Benchmarks are conducted at batch size 1 with \texttt{max\_new\_tokens=1} to isolate the prefill-stage cost. For each configuration, we perform 10 warm-up runs followed by $N=100$ timed iterations, recording the mean ($\mu$) and standard deviation ($\sigma$):
\begin{equation}
\mu = \frac{1}{N} \sum_{i=1}^{N} t_i, \quad \sigma = \sqrt{\frac{1}{N} \sum_{i=1}^{N} (t_i - \mu)^2}
\end{equation}

The end-to-end inference latency $T_{total}$ is decomposed into two primary components: (1) \textbf{ViT Latency} ($T_{ViT}$), the time consumed by the visual encoder and projector; and (2) \textbf{Prefill Latency} ($T_{prefill}$), the language model's forward pass over the multimodal sequence. $T_{prefill}$ is measured by feeding pre-computed embeddings directly into the decoder to isolate language modeling from visual feature extraction.

The total latency is formulated as:
\begin{equation}
T_{total} \approx T_{ViT} + T_{prefill}, \quad \text{where} \quad T_{prefill} = \sum_{l=1}^{D} \text{Layer}(S_l)
\end{equation}
In this formulation, $D$ is the number of decoder layers and $S_l$ denotes the sequence length at layer $l$. This accounts for dynamic token reduction strategies where $S_l$ may vary across layers. While $T_{ViT}$ typically remains constant for a fixed input resolution, the overall efficiency is primarily determined by the adaptive reduction of $S_l$ during the prefill stage.

\section{Additional Experiments}
\subsection{Additional Ablation Studies}

\begin{figure}[t]
    \centering
    \includegraphics[width=0.5\textwidth]{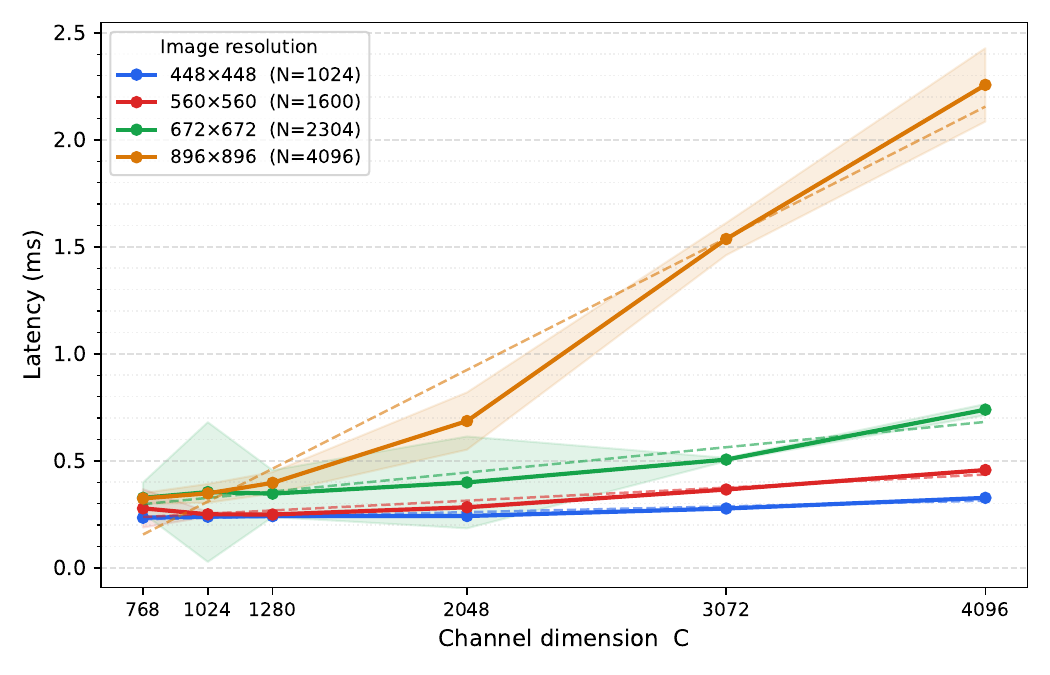}
    \caption{FFT latency as a function of channel dimension $C$ across four spatial resolutions ($N \in \{1024, 1600, 2304, 4096\}$). Latency scales approximately linearly with both $C$ and $N$, remaining in the low-millisecond range across all tested configurations.}
    \label{fig:latency}
\end{figure}

\begin{table}[t]
\centering
\caption{Ablation study on \textbf{Z-Score normalization and temperature} in \ours-Hybird on LLaVA-1.5-7B. Without Z-score, the raw evolution intensity is used directly as the gating factor. Z-score normalization consistently improves performance, and $\tau\!=\!0.1$ achieves the best results across all benchmarks. Excessively large $\tau$ flattens the sigmoid gate toward a uniform weight, degrading token selection.}
\label{tab:ablation_temp}
\begin{adjustbox}{width=0.99\linewidth,center}
\begin{tabular}{l|c|cccc}
\toprule
\textbf{Setting} & \textbf{Temp ($\tau$)} & \textbf{GQA} $\uparrow$ & \textbf{MMB} $\uparrow$ & \textbf{MME} $\uparrow$ & \textbf{POPE} $\uparrow$ \\ \midrule
Vanilla (Full Tokens) & - & 61.9 & 64.7 & 1862 & 85.9 \\ \midrule
\rowcolor[HTML]{F2F2F2}
\multicolumn{6}{l}{\textit{\ours-H Pruning (Retain 192 Tokens, $\downarrow$66.7\%)}} \\ \midrule
No Z-Score & - & 60.5 & 61.9 & 1796 & 82.7 \\
With Z-Score & 0.01 & 60.8 & \textbf{62.7} & 1805 & 83.7 \\
With Z-Score & 0.1 & \textbf{60.9} & \textbf{62.7} & \textbf{1815} & \textbf{83.8} \\
With Z-Score & 1 & 59.9 & \textbf{62.7} & 1781 & 81.6 \\
With Z-Score & 10 & 56.8 & 62.3 & 1751 & 74.6 \\ \bottomrule
\end{tabular}
\end{adjustbox}
\end{table}

\noindent \textbf{FFT Latency Scaling with Channel Dimension.}
Since \ours applies a 2D FFT independently per feature channel, the spectral scoring cost scales as $O(C \cdot N \log N)$, where $C$ is the channel dimension and $N$ is the number of spatial tokens.
The FFT is implemented via \texttt{torch.fft.fft2} (cuFFT backend) over batched channel dimensions in a single kernel call, operating in bfloat16 precision on NVIDIA RTX PRO 6000 Blackwell GPUs.
We benchmark the spectral scoring function in isolation using synthetic inputs, sweeping six channel sizes ($C \in \{768, 1024, 1280, 2048, 3072, 4096\}$) and four spatial resolutions ($N \in \{1024, 1600, 2304, 4096\}$) following the \texttt{patch\_size=14} ViT convention.
As shown in Fig.~\ref{fig:latency}, latency grows approximately linearly with $C$ at each resolution, consistent with the $O(C \cdot N \log N)$ analysis, and remains in the low-millisecond range across all tested configurations.

\noindent \textbf{Z-Score in CLSE-Hybrid.}
In \ours-Hybrid, the per-token evolution intensity $\delta_i$ (defined in the main paper) is Z-score normalized across tokens before being used as a gating factor:
\begin{equation}
\hat{\delta}_i = \frac{\delta_i - \mu_\delta}{\sigma_\delta + \epsilon}, \qquad
e_i = \sigma\!\left(\hat{\delta}_i \,/\, \tau\right),
\end{equation}
where $\tau$ is a temperature parameter. This standardization makes the gate robust to the absolute scale of spectral intensities, which varies across images and models. As shown in Table~\ref{tab:ablation_temp}, omitting Z-score consistently degrades performance, and $\tau=0.1$ achieves the best results by producing a sharp yet smooth discrimination between high- and low-evolution tokens.

\noindent \textbf{Pruning Layer Analysis.}
Table~\ref{tab:ablation_layers} reports the effect of varying the reference layer $L$ and pruning layer $K$ while retaining 192 visual tokens ($\downarrow$66.7\%).
In early layers, visual tokens still preserve their original spatial diversity—semantically active regions (edges, fine structures) evolve rapidly across layers while background tokens remain spectrally static, yielding a highly discriminative CLSE signal; pruning here also maximizes latency savings.
We therefore recommend $K\!\in\!\{1,3\}$ as the sweet spot, balancing accuracy and efficiency.
Mid-range pruning (e.g., $K\!=\!10$) suffers from two compounding issues: repeated layer-wise mixing has already reduced the spectral diversity among visual tokens, weakening the CLSE signal, while most computation has already been incurred.
At very deep layers ($K\!\geq\!15$), visual tokens have become mutually homogenized through extensive layer-wise abstraction—their representations are nearly interchangeable—so accuracy recovers, but all latency benefit is lost.
Comparing reference choices at the same $K$, the adjacent-layer reference ($L\!=\!K\!-\!1$) is generally competitive with or superior to the global reference ($L\!=\!0$), suggesting that local spectral change is a reliable signal.
Fig.~\ref{fig:layers-latency} further visualizes the performance--latency trade-off across pruning layers, confirming that early-layer pruning (layers 1 and 3) achieves the best balance between efficiency and accuracy.

\begin{table}[t]
\centering
\setlength{\tabcolsep}{0pt}
\caption{Ablation study on \textbf{CLSE Layer Selection}. $L$ and $K$ denote the reference and scoring layers used to compute the cross-layer spectral evolution signal; pruning is applied at layer $K$. All settings retain 192 visual tokens ($\downarrow$66.7\%). Our default ($L\!=\!0, K\!=\!1$) achieves competitive performance with the lowest evaluation time. Mid-range pruning (e.g., $K\!=\!10$) causes noticeable degradation, while the adjacent-layer reference ($L\!=\!K\!-\!1$) is generally competitive with or superior to the global reference ($L\!=\!0$) at the same $K$.}
\label{tab:ablation_layers}
\begin{adjustbox}{width=0.99\linewidth,center}
\begin{tabular}{cc|cccc|c}
\toprule
\textbf{Ref.\ Layer ($L$)} & \textbf{Prune Layer ($K$)} & \textbf{GQA} $\uparrow$ & \textbf{MMB} $\uparrow$ & \textbf{MME} $\uparrow$ & \textbf{POPE (F1)} $\uparrow$ & \textbf{Time (s)} $\downarrow$ \\ \midrule
LLaVA-1.5-7B (Full Tokens) & -- & 61.9 & 64.7 & 1862 & 85.9 & 2348 \\ \midrule
\rowcolor[HTML]{F2F2F2}
\multicolumn{7}{l}{\textit{CLSE Pruning (Retain 192 Tokens, $\downarrow$66.7\%)}} \\ \midrule
\multicolumn{7}{l}{\textit{Pruning at Layer 1}} \\
0 & 1 & 61.1 & 62.6 & 1814 & 83.7 & 1691 \\ \midrule
\multicolumn{7}{l}{\textit{Pruning at Layer 3}} \\
0 & 3 & 58.8 & 62.2 & 1758 & 81.9 & 1732 \\
2 & 3 & \textbf{60.4} & \textbf{62.5} & \textbf{1811} & \textbf{83.9} & 1738 \\ \midrule
\multicolumn{7}{l}{\textit{Pruning at Layer 5}} \\
0 & 5 & \textbf{56.4} & \textbf{61.1} & \textbf{1697} & \textbf{74.2} & 1756 \\
4 & 5 & 54.7 & 59.5 & 1653 & 67.4 & 1746 \\ \midrule
\multicolumn{7}{l}{\textit{Pruning at Layer 10}} \\
0 & 10 & 49.4 & 58.0 & 1480 & 42.7 & 1927 \\
9 & 10 & \textbf{54.3} & \textbf{59.3} & \textbf{1637} & \textbf{65.4} & 1932 \\ \midrule
\multicolumn{7}{l}{\textit{Pruning at Layer 15}} \\
0 & 15 & 57.6 & 64.0 & 1845 & 85.7 & 2001 \\
14 & 15 & \textbf{60.8} & \textbf{64.1} & \textbf{1850} & \textbf{85.8} & 2007 \\ \midrule
\multicolumn{7}{l}{\textit{Pruning at Layer 20}} \\
0 & 20 & \textbf{61.6} & \textbf{64.3} & \textbf{1857} & \textbf{86.0} & 2066 \\
19 & 20 & 61.0 & 64.2 & 1853 & \textbf{86.0} & 2075 \\ \bottomrule
\end{tabular}
\end{adjustbox}
\end{table}

\begin{figure*}[t]
  \centering
  \begin{subfigure}[b]{0.49\linewidth}
    \centering
    \includegraphics[width=\linewidth]{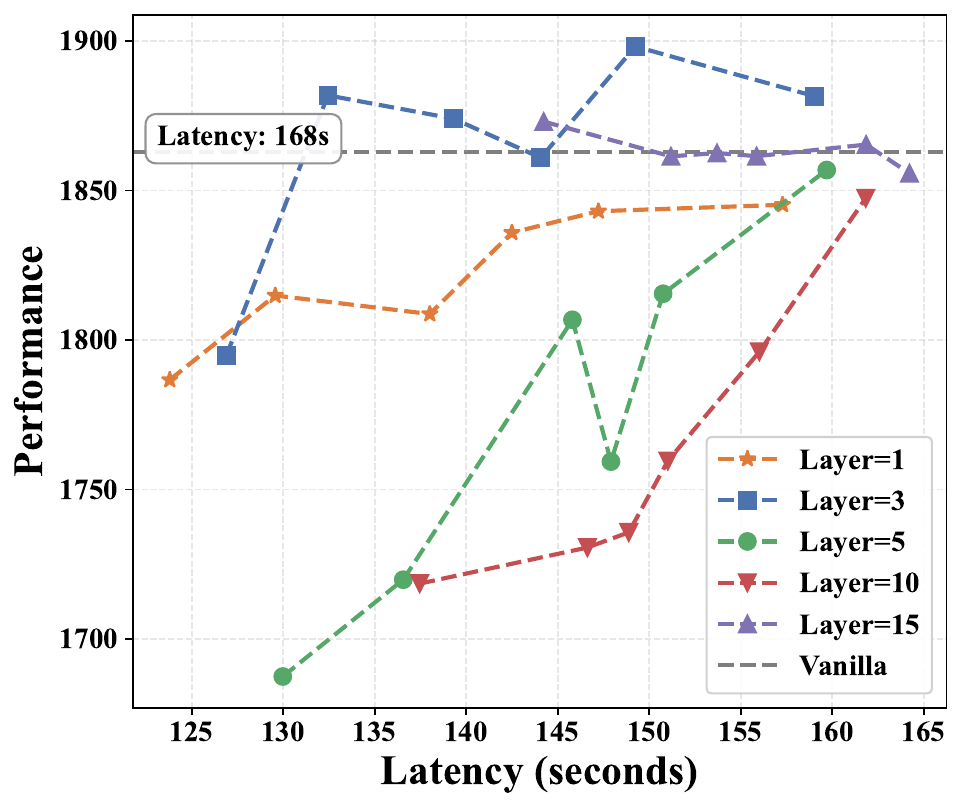}
    \caption{MME}
    \label{fig:layers-latency-mme}
  \end{subfigure}\hfill
  \begin{subfigure}[b]{0.49\linewidth}
    \centering
    \includegraphics[width=\linewidth]{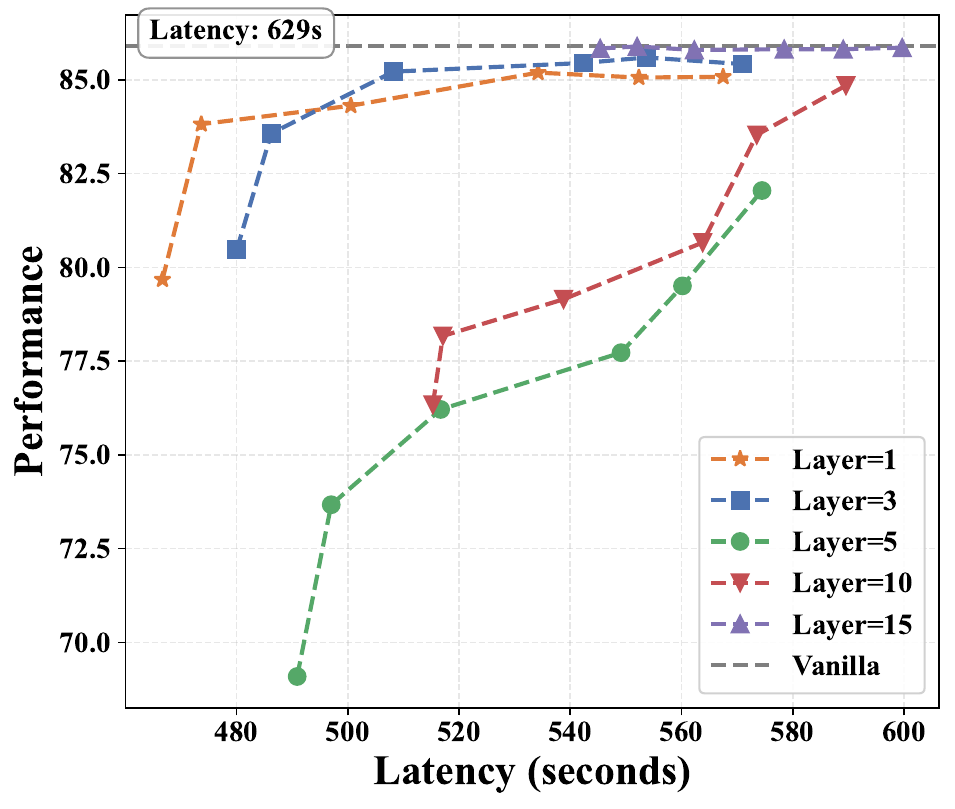}
    \caption{POPE} 
    \label{fig:layers-latency-pope}
  \end{subfigure}
  \caption{Influence of Pruning Layer on Performance. We evaluated five pruning ratios per layer with \ours-Hybrid. While pruning deeper layers maintains performance near or even above the Vanilla, the latency reduction remains marginal. Conversely, pruning shallower layers (e.g., layers 1 and 3) yields competitive results and, notably, outperforms the Vanilla on the MME dataset, offering a superior trade-off between efficiency and accuracy.}
  \label{fig:layers-latency}
\end{figure*}

\subsection{Results on LLaVA-1.5-13B}
To validate generalization to larger model scales, we evaluate \ours on LLaVA-1.5-13B~\cite{liu2024improved} and compare against representative baselines under three compression ratios.
As shown in Table~\ref{app_tab:llava_13B}, \ours achieves the highest average performance retention across all settings, attaining 99.1\% at $\downarrow$66.7\% and maintaining an 8.0 percentage-point lead over the next best method at the aggressive $\downarrow$88.9\% ratio, demonstrating that spectral evolution scoring remains effective as model capacity increases.

\begin{table}[!t]
	\caption{Comparative experiments on LLaVA-1.5-13B.}
    \label{app_tab:llava_13B}
    \centering
    \setlength{\tabcolsep}{2.5pt}
    \renewcommand{\arraystretch}{1.0}
    \resizebox{0.47\textwidth}{!}{
    \begin{tabular}{c | c c c c c c c | >{\centering\arraybackslash}p{1.0cm}}

        \toprule[1.5pt]        \textbf{Method} & \textbf{GQA} & \textbf{MMB} & \textbf{MMB-CN} & \textbf{MME} & \textbf{POPE} & \textbf{SQA} &  \textbf{VQA}$^{\text{Text}}$ & \makecell[c]{\textbf{Avg}.}  \\
        \hline
        \rowcolor{mygray}
        LLaVA-1.5-13B & \multicolumn{8}{c}{\textit{Upper Bound, 576 Tokens} \ $\textbf{(100\%)}$}\\
        \textcolor{gray}{Vanilla} & \textcolor{gray}{63.3} & \textcolor{gray}{68.5} & \textcolor{gray}{62.3} & \textcolor{gray}{1816} & \textcolor{gray}{85.9} & \textcolor{gray}{74.9} & \textcolor{gray}{61.2} & \multirow{1}*{\textcolor{gray}{100\%}} \\
        \hline

        \rowcolor{mygray}
        LLaVA-1.5-13B & \multicolumn{8}{c}{\textit{Retain 192 Tokens} \ $\fg{(\downarrow 66.7\%)}$}\\
        FastV \texttt{\scriptsize{(ECCV24)}} & 59.1 & 54.0 & 51.2 & 1641 & 82.3 & 56.4 & 51.6 & 86.4\% \\

        SparseVLM \texttt{\scriptsize{(ICML25)}} & 58.7 & 67.4 & 61.0 & 1768 & 82.2 & 73.1 & 45.4 & 91.0\% \\
        
        PDrop \texttt{\scriptsize{(CVPR25)}} & 59.6 & \textbf{67.5} & \textbf{62.4} & 1786 & \textbf{85.0} & 74.9 & 59.5 & 98.2\% \\

        FiCoCo-V \texttt{\scriptsize{(AAAI26)}} & 58.1 & 65.2 & 58.3 & 1791 & 81.4 & \textbf{75.1} & 56.2 & 95.1\% \\

        \ours (Ours) & \textbf{62.4} & 67.4 & 60.8 & \textbf{1826} & \textbf{85.0} & 74.7 & \textbf{60.0} & \textbf{99.1\%} \\
        \hline

        \rowcolor{mygray}
        LLaVA-1.5-13B & \multicolumn{8}{c}{\textit{Retain 128 Tokens} \ $\fg{(\downarrow 77.8\%)}$}   \\
        FastV \texttt{\scriptsize{(ECCV24)}} & 57.7 & 57.9 & 48.8 & 1673 & 79.3 & 57.0 & 56.0 & 87.2\%  \\

        SparseVLM \texttt{\scriptsize{(ICML25)}} & 57.9 & 65.8 & 55.8 & 1774 & 81.1 & 69.9 & 49.9 & 94.1\% \\

        PDrop \texttt{\scriptsize{(CVPR25)}} & 57.2 & 65 & 59.2 & 1744 & 82.3 & \textbf{75.4} & \textbf{58.3} & 96.5\% \\

        FiCoCo-V \texttt{\scriptsize{(AAAI26)}} & 56.8 & 62.1 & 55.9 & 1710 & 77.3 & 75.0 & 54.8 & 92.0\% \\

       \ours (Ours) & \textbf{61.6} & \textbf{66.5} & \textbf{59.5} & \textbf{1785} & \textbf{82.9} & 75.0 & \textbf{58.3} & \textbf{98.2\%}   \\
        \hline

        \rowcolor{mygray}
        LLaVA-1.5-13B & \multicolumn{8}{c}{\textit{Retain 64 Tokens} \ $\fg{(\downarrow 88.9\%)}$}\\
        FastV \texttt{\scriptsize{(ECCV24)}} & 53.7 & 50.9 & 42.1 & 1567 & 69.3 & 56.8 & 47.1 & 81.0\% \\
        SparseVLM \texttt{\scriptsize{(ICML25)}} & 50.6 & 61.3 & 54.8 & 1402 & 65.0 & 69.0 & 22.7 & 78.3\% \\
        PDrop \texttt{\scriptsize{(CVPR25)}} & 50.7 & 59.2 & 50.7 & 1441 & 62.6 & \textbf{74.2} & 52.9 & 84.6\% \\

        FiCoCo-V \texttt{\scriptsize{(AAAI26)}} & 54.4 & 55.8 & 49.4 & 1615 & 70.9 & \textbf{74.2} & 52.5 & 86.1\% \\

        \ours (Ours) & \textbf{58.6} & \textbf{63.3} & \textbf{56.7} & \textbf{1728} & \textbf{75.9} & 73.6 & \textbf{54.5} & \textbf{94.1\%} \\        \bottomrule[1.5pt]

	\end{tabular}}
         \vspace{1mm}

    \vspace{2mm}
\end{table}

\subsection{Results on InternVL-2.5-8B}
We additionally evaluate \ours on InternVL-2.5-8B~\cite{chen2024expanding} to assess cross-architecture generalization.
InternVL encodes high-resolution images via dynamic tiling into $n_{\text{sub}}$ sub-tiles ($16{\times}16$ tokens each) plus a 256-token thumbnail.
$\text{\ours}_{\text{local}}$ applies 2D FFT independently per sub-tile, while $\text{\ours}_{\text{global}}$ reassembles sub-tiles into their original spatial layout before a single global FFT.
In both variants, sub-tiles and the thumbnail are scored with separate normalization statistics—since the thumbnail's low-frequency characteristics would otherwise be systematically suppressed under a global mean and standard deviation—and a single global Top-$K$ is then applied across all visual tokens to ensure fair cross-partition selection.
As shown in Table~\ref{app_tab:internvl_8B}, both variants consistently outperform FastV and achieve comparable accuracy to each other, confirming that the CLSE signal is robust to how spatial tokens are partitioned for FFT.
At $\downarrow$88.9\%, both achieve 96.2\% average retention vs.\ 89.4\% for FastV.

\begin{table}[!t]
    \caption{Comparative experiments on InternVL-2.5-8B. $\text{\ours}_{\text{local}}$ applies 2D FFT independently within each $16{\times}16$ sub-tile, while $\text{\ours}_{\text{global}}$ reassembles sub-tiles into the full spatial layout before applying a single 2D FFT, capturing cross-tile global frequency structure. The thumbnail is scored separately in both variants.}
    \label{app_tab:internvl_8B}
    \centering
    \centering
    \resizebox{0.5\textwidth}{!}{
    \begin{tabular}{c | c c c c c c c | >{\centering\arraybackslash}p{1.0cm}}
        \toprule[1.5pt]        \textbf{Method} & \textbf{GQA} & \textbf{MMB} & \textbf{MMB-CN} & \textbf{MME} & \textbf{POPE} & \textbf{SQA} &  \textbf{VQA}$^{\text{Text}}$ & \makecell[c]{\textbf{Avg}.}  \\
        \hline
        \rowcolor{mygray}
        InternVL-2.5-8B & \multicolumn{8}{c}{\textit{Upper Bound, Full Tokens} \ $\textbf{(100\%)}$}\\
        \textcolor{gray}{Vanilla} & \textcolor{gray}{63.1} & \textcolor{gray}{84.7} & \textcolor{gray}{82.6} & \textcolor{gray}{2332} & \textcolor{gray}{90.5} & \textcolor{gray}{98.0} & \textcolor{gray}{76.8} & \multirow{1}*{\textcolor{gray}{100\%}} \\
        \hline

        \rowcolor{mygray}
        InternVL-2.5-8B & \multicolumn{8}{c}{\textit{Token Reduction} \ $\fg{(\downarrow 66.7\%)}$}\\
        FastV \texttt{\scriptsize{(ECCV24)}} & 61.2 & 83.8 & 82.4 & 2252 & 90.4 & 96.4 & 75.5 & 98.4\% \\

        \ours$_{\text{local}}$ & 61.8 & \textbf{84.1} & 82.5 & 2312 & \textbf{90.5} & \textbf{96.6} & \textbf{75.9} & \textbf{99.1\%} \\

        \ours$_{\text{global}}$ & \textbf{61.9} & 83.7 & \textbf{82.6} & \textbf{2325} & \textbf{90.5} & 96.5 & \textbf{75.9} & \textbf{99.1\%} \\
        \hline

        \rowcolor{mygray}
        InternVL-2.5-8B & \multicolumn{8}{c}{\textit{Token Reduction} \ $\fg{(\downarrow 77.8\%)}$}\\
        FastV \texttt{\scriptsize{(ECCV24)}} & 59.6 & 81.1 & 80.5 & 2208 & 89.4 & 95.7 & 73.7 & 96.4\% \\

        \ours$_{\text{local}}$ & 60.7 & \textbf{83.6} & \textbf{82.0} & 2291 & \textbf{90.7} & \textbf{96.4} & 74.9 & \textbf{98.4\%} \\

        \ours$_{\text{global}}$ & \textbf{60.9} & 83.5 & 81.8 & \textbf{2293} & 90.6 & \textbf{96.4} & \textbf{75.0} & \textbf{98.4\%} \\
        \hline

        \rowcolor{mygray}
        InternVL-2.5-8B & \multicolumn{8}{c}{\textit{Token Reduction} \ $\fg{(\downarrow 88.9\%)}$}\\
        FastV \texttt{\scriptsize{(ECCV24)}} & 54.1 & 73.9 & 74.4 & 2052 & 85.2 & 92.5 & 66.4 & 89.4\% \\

        \ours$_{\text{local}}$ & 59.4 & 81.4 & 80.3 & 2201 & \textbf{89.7} & \textbf{95.8} & \textbf{72.5} & \textbf{96.2\%} \\

        \ours$_{\text{global}}$ & \textbf{59.6} & \textbf{81.7} & \textbf{80.5} & \textbf{2204} & 89.6 & 95.7 & 71.9 & \textbf{96.2\%} \\
        \bottomrule[1.5pt]
    \end{tabular}}
    \vspace{1mm}

    \vspace{2mm}
\end{table}

\subsection{Results on Qwen2-VL-72B}
We further evaluate \ours on Qwen2-VL-72B~\cite{wang2024qwen2vl} with 4-bit weight quantization to assess scalability to very large models.
As shown in Table~\ref{tab:qwen2_vl_72B}, \ours consistently outperforms both FastV and PDrop across all three compression ratios, with a performance retention gap over FastV that widens monotonically from 2.0 to 3.7 to 7.8 percentage points as pruning becomes more aggressive ($\downarrow$66.7\% $\to$ $\downarrow$88.9\%).
This demonstrates that cross-layer spectral evolution scoring remains effective even at 72B scale under quantized weights.

\renewcommand{\multirowsetup}{\centering}
\definecolor{mygray}{gray}{.92}
\definecolor{ForestGreen}{RGB}{34,139,34}
\definecolor{Forestred}{RGB}{220,50,50}
\begin{table}[t]
 \caption{Comparative experiments on Qwen2-VL-72B.}
    \centering
    \setlength{\tabcolsep}{2.5pt}
    \renewcommand{\arraystretch}{1.25}
    \footnotesize
    \centering
    \scalebox{0.9}{
    \begin{tabular}{c | c c c c c c | >{\centering\arraybackslash}p{1.0cm}}
        \toprule[1.5pt]

        \textbf{Method}& \textbf{GQA} & \textbf{MMB} & \textbf{MME} & \textbf{POPE} & \textbf{SQA} & \textbf{VQA}$^{\text{Text}}$ & \makecell[c]{\textbf{Avg}.} \\
        \hline        \rowcolor{mygray}
        Qwen2-VL-72B & \multicolumn{7}{c}{\textit{Upper Bound, Full Tokens} \ $\textbf{(100\%)}$}\\
        \textcolor{gray}{Vanilla} & \textcolor{gray}{64.8} & \textcolor{gray}{85.0} & \textcolor{gray}{2476} & \textcolor{gray}{87.5} & \textcolor{gray}{92.5} & \textcolor{gray}{82.0} & \textcolor{gray}{100\%} \\

        \hline        \rowcolor{mygray}
        Qwen2-VL-72B & \multicolumn{7}{c}{\textit{Token Reduction} \ $\fg{(\downarrow 66.7\%)}$}\\

        FastV \texttt{\scriptsize{(ECCV24)}} & 59.6 & 79.0 & 2314 & 82.8 & 89.9 & \textbf{79.1} & 94.4\% \\
        PDrop \texttt{\scriptsize{(CVPR25)}} & 60.9 & 81.3 & 2273 & 84.9 & 90.5 & 78.9 & 95.4\% \\
        \ours~(Ours) & \textbf{61.4} & \textbf{82.4} & \textbf{2340} & \textbf{85.6} & \textbf{91.3} & 78.5 & \textbf{96.4\%} \\
        \hline        \rowcolor{mygray}
        Qwen2-VL-72B & \multicolumn{7}{c}{\textit{Token Reduction} \ $\fg{(\downarrow 77.8\%)}$}\\

        FastV \texttt{\scriptsize{(ECCV24)}} & 57.2 & 76.9 & 2173 & 78.0 & 88.7 & 76.6 & 90.8\% \\
        PDrop \texttt{\scriptsize{(CVPR25)}} & 58.5 & 77.8 & 2178 & 82.0 & 89.2 & \textbf{76.7} & 92.2\% \\
        \ours~(Ours) & \textbf{60.1} & \textbf{79.3} & \textbf{2309} & \textbf{84.7} & \textbf{90.6} & 76.0 & \textbf{94.5\%} \\
        \hline        \rowcolor{mygray}
        Qwen2-VL-72B & \multicolumn{7}{c}{\textit{Token Reduction} \ $\fg{(\downarrow 88.9\%)}$}\\
        FastV \texttt{\scriptsize{(ECCV24)}} & 53.0 & 66.0 & 1869 & 71.3 & 86.1 & 68.8 & 82.2\% \\
        PDrop \texttt{\scriptsize{(CVPR25)}} & 54.6 & 67.5 & 2057 & 75.6 & 87.0 & 70.0 & 85.4\% \\
        \ours~(Ours) & \textbf{57.2} & \textbf{75.3} & \textbf{2143} & \textbf{81.4} & \textbf{88.6} & \textbf{72.1} & \textbf{90.0\%} \\
        \bottomrule[1.5pt]
    \end{tabular}}

    \label{tab:qwen2_vl_72B}
\end{table}

\section{Theoretical Analysis}
We next show that the proposed pruning rule is not merely heuristic. 
Under a fixed token budget, selecting tokens by the proposed cross-layer spectral evolution score maximizes the preserved amount of informative structural dynamics. 
Moreover, under a mild downstream smoothness assumption, the same rule minimizes an upper bound on the perturbation induced by pruning.

\noindent \textbf{Setup.}
Let ${\mathcal I}^{(\ell)} \in \mathbb{R}^{N}$ denote the calibrated evolution scores defined in Eq.~(9), where $N=H\!\cdot\!W$ is the number of visual tokens.
For notational simplicity, we define the nonnegative token score as:
\begin{equation}
s_i := {\mathcal I}^{(\ell)}_i, \qquad i\in\{1,\dots,N\}.
\end{equation}
For any retained token set $\Omega \subseteq \{1,\dots,N\}$ with $|\Omega|=K$, let
$P_{\Omega}(\mathbf X^\ell)$
denote the pruned token sequence that keeps tokens indexed by $\Omega$ and discards the others.
We further define the \emph{retained spectral evolution mass} as:
\begin{equation}
\mathcal Q(\Omega) := \sum_{i\in\Omega} s_i ,
\end{equation}
which measures the total amount of cross-layer structural evolution preserved after pruning.

\begin{proposition}[Optimal preservation under a cardinality budget]
\label{prop:topk_optimal}
Let $\Omega_K^\star$ be the index set of the top-$K$ entries of $\{s_i\}_{i=1}^N$. Then
\begin{equation}
\Omega_K^\star
\in
\arg\max_{|\Omega|=K} \mathcal Q(\Omega).
\end{equation}
Equivalently,
\begin{equation}
\Omega_K^\star
\in
\arg\min_{|\Omega|=K} \sum_{i\notin\Omega} s_i .
\end{equation}
\end{proposition}

\noindent
Proposition~\ref{prop:topk_optimal} states that, among all $K$-token subsets, the proposed top-$K$ rule preserves the maximum amount of cross-layer spectral evolution, or equivalently discards the minimum amount of structural evolution mass.

\paragraph{Why spectral evolution matters.}
Unlike magnitude-based criteria, our score is computed from high-pass filtered residual maps:
\begin{equation}
\mathbf S_\ell = \mathrm{Avg}_d\!\left(
\left|
\mathcal F^{-1}\!\big(
\mathrm{ifftshift}(\mathrm{fftshift}(\mathcal{F}(\mathbf{X}^{\ell}_{2\mathrm{D}}))\odot \mathbf M)
\big)
\right|
\right),
\end{equation}
and therefore quantifies \emph{band-limited structural activity} rather than raw embedding norm.
Large $s_i$ indicates that token $i$ undergoes substantial redistribution of local high-frequency content across adjacent layers, which is precisely the regime where text-conditioned visual reasoning actively reshapes token representations.
Thus, preserving large-$s_i$ tokens preserves the most dynamically updated local structures.
To relate this surrogate quantity to downstream fidelity, we consider the remaining decoder after pruning.

\noindent \textbf{Assumption 1} \textit{(Downstream smoothness).}
Let $g(\cdot)$ denote the subsequent decoder mapping from the pruned visual-token sequence to the final hidden representation or output logits.
We assume $g$ is locally Lipschitz:
\begin{equation}
\|g(\mathbf U)-g(\mathbf V)\|_2
\le
L\|\mathbf U-\mathbf V\|_F,
\end{equation}
for some constant $L>0$ and all $\mathbf U,\mathbf V$ in a local neighborhood of interest.

\noindent \textbf{Assumption 2} \textit{(Saliency alignment).}
We assume that the downstream influence of each token is upper bounded by its spectral evolution score.
Formally, let $\mathbf J_i$ denote the Jacobian block of $g$ with respect to token $\mathbf x_i^\ell$. Then there exists $\alpha>0$ such that:
\begin{equation}
\|\mathbf J_i \mathbf x_i^\ell\|_2 \le \alpha s_i,
\qquad \forall i\in\{1,\dots,N\}.
\end{equation}
This assumption formalizes the inductive bias of our method: tokens with weak cross-layer spectral evolution have limited effect on the downstream computation, whereas tokens with strong evolution are more influential.

\begin{theorem}[Top-$K$ spectral evolution minimizes an output perturbation bound]
\label{thm:perturbation}
Under Assumptions 1--2, for any retained set $\Omega$ with $|\Omega|=K$, the pruning-induced perturbation satisfies:
\begin{equation}
\|g(\mathbf X^\ell)-g(P_\Omega(\mathbf X^\ell))\|_2
\le
\alpha \sum_{i\notin\Omega} s_i .
\end{equation}
Consequently, the top-$K$ set $\Omega_K^\star$ from Proposition~\ref{prop:topk_optimal} minimizes the above upper bound among all subsets of size $K$:
\begin{equation}
\|g(\mathbf X^\ell)-g(P_{\Omega_K^\star}(\mathbf X^\ell))\|_2
\le
\alpha \min_{|\Omega|=K}\sum_{i\notin\Omega} s_i .
\end{equation}
\end{theorem}

\begin{proof}
The optimality of $\Omega_K^\star$ in Proposition~\ref{prop:topk_optimal} follows directly from sorting nonnegative scores: among all subsets of cardinality $K$, the sum of retained scores is maximized by selecting the $K$ largest entries.

We now prove the perturbation bound.
Let $\Delta_\Omega := \mathbf X^\ell - P_\Omega(\mathbf X^\ell)$ denote the perturbation introduced by pruning.
By a first-order expansion of $g$ around $\mathbf X^\ell$, we have
\begin{equation}
g(\mathbf X^\ell)-g(P_\Omega(\mathbf X^\ell))
=
g(\mathbf X^\ell)-g(\mathbf X^\ell-\Delta_\Omega)
\approx
\sum_{i\notin\Omega}\mathbf J_i \mathbf x_i^\ell .
\end{equation}
Applying the triangle inequality yields
\begin{equation}
\|g(\mathbf X^\ell)-g(P_\Omega(\mathbf X^\ell))\|_2
\le
\sum_{i\notin\Omega}\|\mathbf J_i \mathbf x_i^\ell\|_2 .
\end{equation}
By Assumption 2,
\begin{equation}
\|g(\mathbf X^\ell)-g(P_\Omega(\mathbf X^\ell))\|_2
\le
\alpha \sum_{i\notin\Omega} s_i .
\end{equation}
Finally, Proposition~\ref{prop:topk_optimal} implies that the top-$K$ subset minimizes the discarded score mass $\sum_{i\notin\Omega}s_i$, and therefore minimizes the above upper bound as well.
\end{proof}
\begin{figure}[t]
  \centering
  \includegraphics[width=1.0\linewidth]{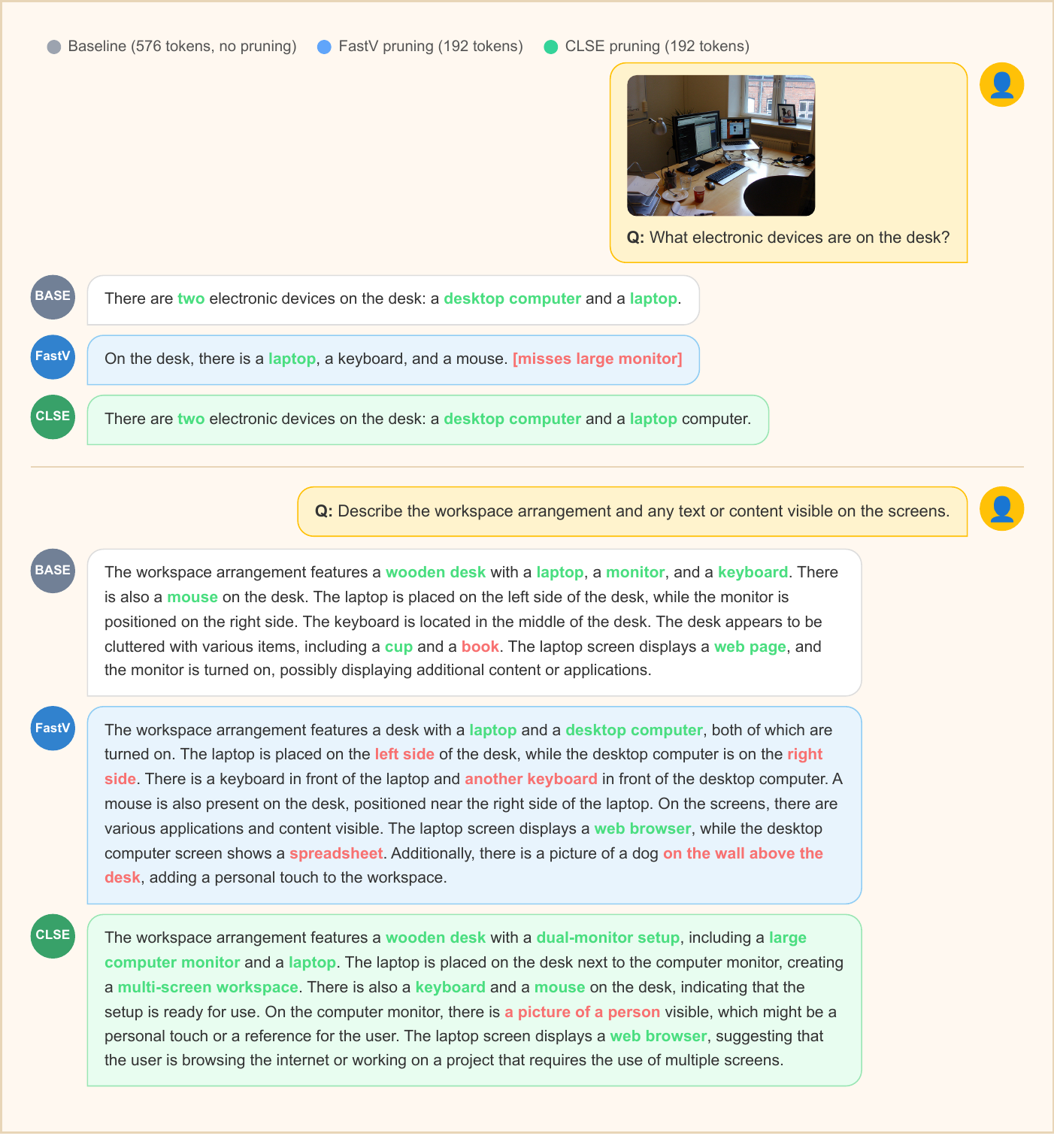}
  \caption{Multi-turn dialogue example on LLaVA-1.5-7B. The user asks about electronic devices on a desk, followed by a request for a detailed workspace description.}
  \label{fig:multiturn_computer}
\end{figure}

\paragraph{Interpretation.}
Theorem~\ref{thm:perturbation} provides a principled justification for our pruning rule.
The proposed criterion does not preserve tokens with merely large activations; instead, it preserves tokens with the strongest \emph{relative cross-layer structural evolution} after low-frequency suppression.
Under a fixed budget, this rule retains the maximum amount of informative spectral dynamics and minimizes a certified upper bound on pruning-induced distortion.
This explains why \ours\ is particularly effective in early decoding, where pruning must be aggressive yet reliable.

\paragraph{Remark.}
Our guarantee is stated with respect to a \emph{structure-preserving surrogate} induced by spectral evolution, rather than a task-specific semantic objective.
This is appropriate for training-free pruning, where explicit task supervision is unavailable at pruning time.
Empirically, the strong alignment between spectral evolution and downstream utility is validated by the ablations in Sec. 4.3.

\begin{figure}[t]
  \centering
  \includegraphics[width=1.0\linewidth]{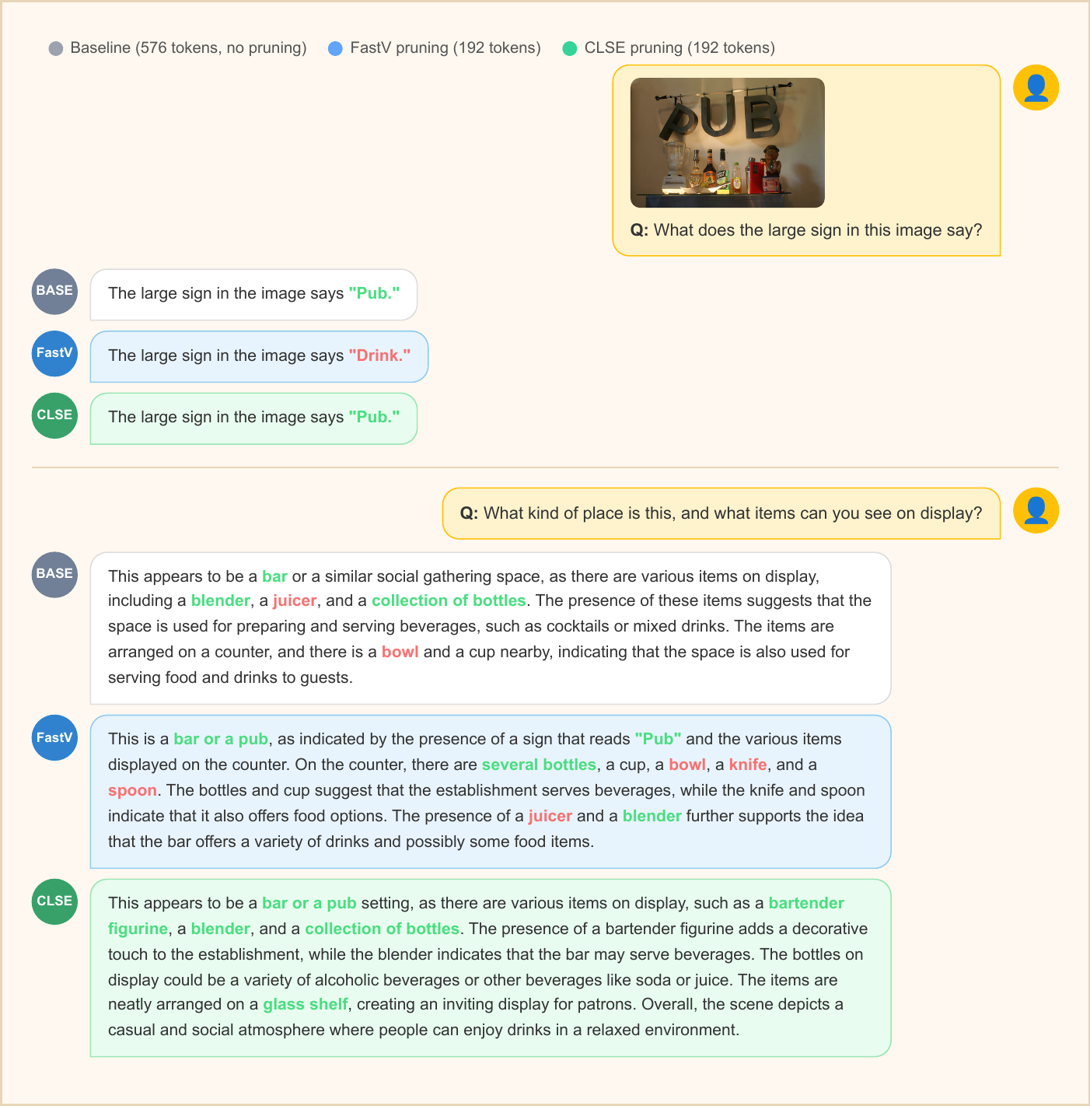}
  \caption{Multi-turn dialogue example on LLaVA-1.5-7B. The user asks about a sign in a pub image, followed by a question about the type of place and items on display.}
  \label{fig:multiturn_pub}
\end{figure}

\section{Qualitative Multi-Turn Dialogue Examples}
\label{app:multiturn}

To illustrate the impact of token pruning on response quality in open-ended dialogue, we compare three settings on LLaVA-1.5-7B: (1)~\textbf{Base}, which processes all 576 visual tokens with no pruning; (2)~\textbf{FastV}~\cite{chen2024image}, which prunes down to 192 visual tokens; and (3)~\textbf{\ours}, which also retains 192 tokens using our spectral evolution-guided criterion.

Figures~\ref{fig:multiturn_computer} and~\ref{fig:multiturn_pub} each show a two-turn conversation over a single image.
The first turn asks a short, factual question, while the second turn requires a more detailed description of the scene.
Despite using only $192/576 \approx 33.4\%$ of the visual tokens, \ours\ consistently produces responses that match the Base model in accuracy and detail.
FastV, by contrast, occasionally drops or misidentifies salient objects (e.g., misses the large monitor in Fig.~\ref{fig:multiturn_computer}; misreads the sign as ``Drink'' instead of ``Pub'' in Fig.~\ref{fig:multiturn_pub}), highlighting the benefit of retaining the most informative tokens as identified by spectral evolution.

\endgroup
\end{document}

%% file: table/image_understanding_llavanext.tex
\renewcommand{\multirowsetup}{\centering}
\definecolor{mygray}{gray}{.92}
\definecolor{ForestGreen}{RGB}{34,139,34}
\definecolor{Forestred}{RGB}{220,50,50}
\begin{table*}[!t]
	\caption{Comparative experiments are performed on LLaVA-Next-7B.}
    \centering
    \centering
\begin{adjustbox}{width=\textwidth,center}
    \begin{tabular}{c | c c c c c c c c c| >{\centering\arraybackslash}p{1.0cm}}
        \toprule[1.5pt]
        \textbf{Method} & \textbf{GQA} & \textbf{MMB} & \textbf{MMB-CN} & \textbf{MME} & \textbf{POPE} & \textbf{SQA} & \textbf{VQA}$^{\text{Text}}$ & \textbf{VizWiz} & \textbf{OCRBench} & \makecell[c]{\textbf{Avg}.}\\
        \hline
        \rowcolor{mygray}
        LLaVA-Next-7B & \multicolumn{10}{c}{\textit{Upper Bound, 2880  Tokens} \ $\textbf{(100\%)}$}\\
         \textcolor{gray}{Vanilla} & \textcolor{gray}{64.2} & \textcolor{gray}{67.4} & \textcolor{gray}{60.6} & \textcolor{gray}{1851} & \textcolor{gray}{86.5} & \textcolor{gray}{70.1} & \textcolor{gray}{64.9} & \textcolor{gray}{57.6} & \textcolor{gray}{51.7} &  \textcolor{gray}{100.0\%} \\
          \hline
       \rowcolor{mygray}
        LLaVA-Next-7B & \multicolumn{10}{c}{\textit{Retain 320 Tokens} \ $\fg{(\downarrow 88.9\%)}$} \\

        FastV \texttt{\scriptsize{(ECCV24)}} & 55.9 & 61.6 & 51.9 & 1661 & 71.7 & 62.8 & 55.7 & 53.1 & 37.4 & 86.3\% \\
        
        HiRED \texttt{\scriptsize{(AAAI25)}} & 59.3 & 64.2 & 55.9 & 1690 & 83.3 & 66.7 & 58.8 & 54.2 & 40.4 & 91.7\% \\
        
        PruMerge \texttt{\scriptsize{(ICCV2025)}} & 53.6 & 61.3 & 55.3 & 1534 & 60.8 & 66.4 & 50.6 & 54.0 & 14.6 & 79.2\% \\

       SparseVLM \texttt{\scriptsize{(ICML25)}} & 56.1 & 60.6 & 54.5 & 1533 & 82.4 & 66.1 & 58.4 & 52.0 & 27.0 & 85.8\%  \\

       PDrop \texttt{\scriptsize{(CVPR25)}} & 56.4 & 63.4 & 56.2 & 1663 & 77.6 & 67.5 & 54.4 & 54.1 & 25.9 & 86.5\% \\


       

      \rowcolor{cyan!7}  \ours (Ours)& \textcolor{MidnightBlue}{\textbf{62.6}} & \textcolor{MidnightBlue}{\textbf{67.1}} & \textcolor{MidnightBlue}{\textbf{58.3}} & \textcolor{MidnightBlue}{\textbf{1794}} & \textcolor{MidnightBlue}{\textbf{84.4}} & \textcolor{MidnightBlue}{\textbf{67.9}} & \textcolor{MidnightBlue}{\textbf{58.9}} & \textcolor{MidnightBlue}{\textbf{56.1}} & \textcolor{MidnightBlue}{\textbf{41.4}} & \textcolor{MidnightBlue}{\textbf{94.7\%}}  \\
        \bottomrule[1.5pt]
\end{tabular}
\end{adjustbox}
    \label{tab:main_1_6}

\end{table*}

%% file: table/fft_2d_vs_3d_ablation.tex
\begin{table}[!t]
\caption{Ablation on \textbf{2D per-frame FFT vs.\ 3D spatiotemporal FFT} for video token scoring on Video-LLaVA-7B, retaining 194 tokens ($\downarrow$90.5\%). $\text{CLSE}_{2\text{df}}$ applies 2D FFT independently to each sampled frame, while $\text{CLSE}_{3\text{d}}$ applies a single 3D FFT over the full spatiotemporal volume. Per-frame 2D FFT consistently and substantially outperforms 3D FFT across all benchmarks and evaluators.}
\centering
\begin{adjustbox}{width=1.0\columnwidth,center}
\renewcommand{\arraystretch}{0.8}
\begin{tabular}{@{}lc cccc cccc cc@{}}
\toprule[1.5pt]
\multirow{2}{*}{\textbf{Method}} & \multirow{2}{*}{\textbf{Eval}}
  & \multicolumn{2}{c}{\textbf{TGIF}}
  & \multicolumn{2}{c}{\textbf{MSVD}}
  & \multicolumn{2}{c}{\textbf{MSRVTT}}
  & \multicolumn{2}{c}{\textbf{ActivityNet}}
  & \multicolumn{2}{c}{\textbf{Avg.}} \\
\cmidrule(lr){3-4}\cmidrule(lr){5-6}\cmidrule(lr){7-8}\cmidrule(lr){9-10}\cmidrule(lr){11-12}
 & & Acc. & Score & Acc. & Score & Acc. & Score & Acc. & Score & Acc. & Score \\
\midrule
\multirow{2}{*}{Video-LLaVA-7B}
  & {\small\textcolor{gray}{GPT}} & \textcolor{gray}{11.3} & \textcolor{gray}{1.19} & \textcolor{gray}{58.9} & \textcolor{gray}{3.31} & \textcolor{gray}{42.7} & \textcolor{gray}{2.69} & \textcolor{gray}{42.0} & \textcolor{gray}{2.20} & \textcolor{gray}{38.7} & \textcolor{gray}{2.35} \\
  & {\small\textcolor{gray}{Gem}} & \textcolor{gray}{14.2} & \textcolor{gray}{0.77} & \textcolor{gray}{57.3} & \textcolor{gray}{2.93} & \textcolor{gray}{47.5} & \textcolor{gray}{2.32} & \textcolor{gray}{42.5} & \textcolor{gray}{2.12} & \textcolor{gray}{40.4} & \textcolor{gray}{2.04} \\
\midrule
\rowcolor{gray!7}\multicolumn{12}{c}{\textit{\footnotesize Retain 194 Tokens}} \\[-2pt]
\midrule
\multirow{2}{*}{$\text{CLSE}_{3\text{d}}$}
  & {\small\textcolor{gray}{GPT}} & 3.0 & 1.08 & 36.7 & 2.37 & 31.4 & 2.16 & 35.1 & 1.87 & 26.5 & 1.87 \\
  & {\small\textcolor{gray}{Gem}} & 3.3 & 0.18 & 36.4 & 1.89 & 37.0 & 1.80 & 36.3 & 1.81 & 28.2 & 1.42 \\
\midrule
\multirow{2}{*}{$\text{CLSE}_{2\text{df}}$}
  & {\small\textcolor{gray}{GPT}} & \textbf{9.6} & \textbf{1.06} & \textbf{54.1} & \textbf{3.11} & \textbf{39.3} & \textbf{2.56} & \textbf{39.9} & \textbf{2.03} & \textbf{35.7} & \textbf{2.19} \\
  & {\small\textcolor{gray}{Gem}} & \textbf{12.9} & \textbf{0.69} & \textbf{56.3} & \textbf{2.88} & \textbf{47.7} & \textbf{2.27} & \textbf{39.9} & \textbf{1.99} & \textbf{39.2} & \textbf{1.95} \\
\bottomrule[1.5pt]
\end{tabular}
\end{adjustbox}
\label{tab:ablation_2d_vs_3d}
\end{table}